\documentclass{article}
\usepackage{arxiv}
\newcommand\alumina{$\alpha$-Al\textsubscript{2}O\textsubscript{3}}

\usepackage[strings]{underscore}
\usepackage[utf8]{inputenc} 
\usepackage[T1]{fontenc}    
\usepackage{hyperref}       
\usepackage{url}            
\usepackage{booktabs}       
\usepackage{amsfonts}       
\usepackage{nicefrac}       
\usepackage{microtype}      
\usepackage{graphicx}
\usepackage[numbers,square]{natbib}
\usepackage{appendix}
\usepackage{amssymb}
\usepackage{amsmath}
\usepackage{caption}
\usepackage[capitalise]{cleveref}
\usepackage{csquotes}
\usepackage{subcaption}
\usepackage{adjustbox}
\usepackage{underscore}
\usepackage{multirow}
\usepackage[dvipsnames]{xcolor}
\usepackage{graphicx}
\usepackage{siunitx}
\usepackage{xcolor}

\title{Implementing NLPs in Industrial Process modeling: Addressing categorical variables}

\author{ \href{https://orcid.org/0000-0002-5229-4157}{\includegraphics[scale=0.06]{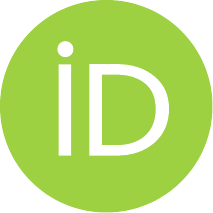}\hspace{1mm}Eleni D.~Koronaki}\thanks{Corresponding author: \texttt{eleni.koronaki@uni.lu}} \\
	Faculty of Science, Technology and Medicine\\
	University of Luxembourg\\
	Esch-sur-Alzette, L-4364, Luxembourg\\
	\texttt{eleni.koronaki@uni.lu}
	\And
\href{https://orcid.org/0009-0002-4531-7960}{\includegraphics[scale=0.06]{orcid.pdf}\hspace{1mm}Geremy Loachamin Suntaxi\thanks{Authors also affiliated with the School of Chemical Engineering, National Technical University of Athens, Zographos Campus, 15780, Attiki, Greece}}\\
	Faculty of Science, Technology and Medicine\\
	University of Luxembourg\\
	Esch-sur-Alzette, L-4364, Luxembourg\\
    \texttt{geremy.loachamin@uni.lu}
    \And
  \href{https://orcid.org/0000-0002-3136-1402}{\includegraphics[scale=0.06]{orcid.pdf}\hspace{1mm}Paris Papavasileiou\footnotemark[1]}\\
	Faculty of Science, Technology and Medicine\\
	University of Luxembourg\\
	Esch-sur-Alzette, L-4364, Luxembourg\\
    \texttt{paris.papavasileiou@uni.lu}
    \And
	\href{https://orcid.org/0000-0003-2272-2584}
    {\includegraphics[scale=0.06]{orcid.pdf}\hspace{1mm}Dimitrios G.~Giovanis} \\
	Department of Civil \& Systems Engineering \\
    Johns Hopkins University\\
    Baltimore, MD 21218, USA\\
	\texttt{dgiovan1@jhu.edu}\\
    \And
	Martin Kathrein \\
	CERATIZIT Luxembourg S.à r.l.\\
    Mamer, L-8232, Luxembourg\\
	\texttt{Martin.Kathrein@ceratizit.com} 
    \And
    Christoph Czettl \\
	CERATIZIT Austria GmbH\\
    Reutte, A-6600, Austria\\
	\texttt{Christoph.Czettl@ceratizit.com} 
    \And
	Andreas G. Boudouvis \\
	School of Chemical Engineering, \\ National Technical University of Athens\\
    Zographos, 15780, Greece\\
	\texttt{boudouvi@chemeng.ntua.gr} 
    \And
	\href{https://orcid.org/0000-0001-7622-2193}{\includegraphics[scale=0.06]{orcid.pdf}\hspace{1mm}St\'{e}phane P.A.~Bordas} \\
	Faculty of Science, Technology and Medicine\\
	University of Luxembourg\\
	Esch-sur-Alzette, L-4364, Luxembourg \\
	\texttt{stephane.bordas@uni.lu}
}

\begin{document}
\maketitle
\begin{abstract}
Important variables of processes are often categorical, i.e. names or labels representing, e.g. categories of inputs, or types of reactors or a sequence of steps. In this work, we use Natural Language Processing Models to derive embeddings of such inputs that represent their actual meaning, or reflect the ``distances" between categories, i.e. how similar or dissimilar they are. This is a marked difference from the current standard practice of using binary, or one-hot encoding to replace categorical variables with sequences of ones and zeros. Combined with dimensionality reduction techniques, either linear such as Principal Component Analysis, or nonlinear such as Uniform Manifold Approximation and Projection, the proposed approach leads to a \textit{meaningful}, low-dimensional feature space. The significance of obtaining meaningful embeddings is illustrated in the context of an industrial coating process for cutting tools that includes both numerical and categorical inputs. In this industrial process, subject matter expertise suggests that the categorical inputs are critical for determining the final outcome but this cannot be taken into account with the current state-of-the-art. The proposed approach enables feature importance which is a marked improvement compared to the current state-of-the-art in the encoding of categorical variables. The proposed approach is not limited to the case-study presented here and is suitable for applications with similar mix of categorical and numerical critical inputs. 
\end{abstract}


\section{Introduction}
Machine learning is currently extensively used in process modeling, design and engineering for a wide array of tasks, ranging from prediction with regression methods \citep{azadiHybridDynamicModel2022,malleyPredictabilityMechanicalBehavior2022}, clustering and classification \citep{kimImbalancedClassificationManufacturing2018,saqlainVotingEnsembleClassifier2019, penumuruIdentificationClassificationMaterials2020}, the development of surrogate models of digital twins \citep{wang2020minilmdeepselfattentiondistillation,chakrabortyRoleSurrogateModels2021,spencer2021investigation} and optimization and control \citep{dornheimModelFreeAdaptiveOptimal2020,humfeldMachineLearningFramework2021}.

When dealing with industrial/production data, it is often the case that data are not numerical values, e.g. temperature time-series, flow-rates, pressure etc., but rather involve categorical variables, i.e. categories described by an assigned name. Examples include serial numbers of reactors, working-names of products or even an entire sequence of steps represented by a single name. One of the most popular methods to address this issue is one-hot encoding \cite{Nelson1955, scikitlearn, murphy2022probabilistic}, which replaces categorical variables with several columns of dummy variables, commonly a sequence of zeros and ones. By replacing categorical variables with numerical ones, it is possible to use them as features in various machine learning algorithms. However, by doing so, the resulting representation is agnostic of the actual meaning of the variables, and it is unable to reflect the actual "distance" (i.e. relative similarity or dissimilarity) between the different categories. For these reasons, it is an obstacle for feature importance and sensitivity analysis, uncertainty quantification, and/or optimization, diminishing the opportunity to create insights from the data and increase understanding of the physical process at hand. 


Following their rise to popularity, a lot of interest has fallen in Natural Language Processing (NLP) models and their implementation \citep{NLPLee, shanahanTalkingLargeLanguage2024}, because these models can generate embeddings that represent categorical variables as dense vectors of real numbers, which can be incorporated into other computational tasks. Specifically in chemical and process engineering, the prospects of NLP, for materials characterization
\citep{tsaiExploringUseLarge2023} and text mining \cite{KUMAR202290, SHAO2024105636, Olivetti2020} have been examined. In \citep{8eedc13120bd4246ad2fba92e65a5f11}, the use of an NLP approach in the context of vapor deposition process is proposed, combining a systems-based approach with NLP to distinguish essential mechanisms and efficiently extract causal knowledge from extensive literature. Furthermore, \cite{ruan2024languagemodelingtabulardata} systematically examines the development of language modeling for tabular data, covering data structures and types, key datasets and evaluation tasks, modeling techniques and the evolution of pre-trained models.
However, despite the widespread use of NLP for text embedding, exemplified by S-Bert \citep{BERT}, to the best of our knowledge, the impact of using Language models to encode categorical variables in the development of predictive models has not been explored.

With this study, we aim to explore the potential benefits of implementing NLPs to derive meaningful embeddings of categorical data, using as inputs short textual descriptions of what the different categories represent. These embeddings are then compressed using either linear or nonlinear dimensionality reduction methods and used as inputs to tree-based regression models. Actual production data are used in this study, stemming from an industrial chemical vapor deposition reactor used for the production of wear-resisting coatings on cutting tools, called ``inserts". This process combines competing physical and chemical mechanisms in very complex and ever-changing geometrical set-ups. There are no inline measurements and therefore there is no possibility to monitor the process as it happens and intervene with control actions. Coating thickness measurements are collected ex-situ, \textit{after} the process is finished, providing information about the quality of the product. 

Aiming to develop a data-driven predictive model \cite{papavasileiou2024integrating} for the quality characteristics of the coating, a tree-based regression algorithm is trained using several numerical and categorical inputs. The categorical inputs include the insert (cutting tool) name, a series of letters and numbers, representing an object with an ISO-specified shape and size. In what we call the ``original"  state-of-the-art (SotA) treatment of the data, the entire name is encoded with one-hot encoding, which is oblivious of the meaning of each component of the name. This poses a restriction on the amount of information that may be extracted from the data-driven model. As an example, it is not possible to establish if and what effect the distribution of inserts in the reactors, their placement, in general, has on the output, i.e. the quality of the product.

In this work, several models, pre-trained an un-trained, are compared and subsequent feature importance analysis, using Shapley values, reveals that critical process inputs; physically meaningful and useful conclusions arise when using the proposed embeddings in contrast to binary encoding.

The development of the proposed tool leads to improved understanding of the process set up since it paves the way for sensitivity analysis and uncertainty quantification. Furthermore, it allows the incorporation into datasets of pieces of production information that are maintained in the form of notes, short texts, in the production diaries. This information is often very important and is assessed by experts in the production line but not in a systematic way.

\section{Process description}
\label{sec:process-overview}
\begin{figure}[ht]
\captionsetup[subfigure]{justification=centering}
\centering
\begin{subfigure}[b]{0.4\textwidth}
   \includegraphics[width=1\textwidth]{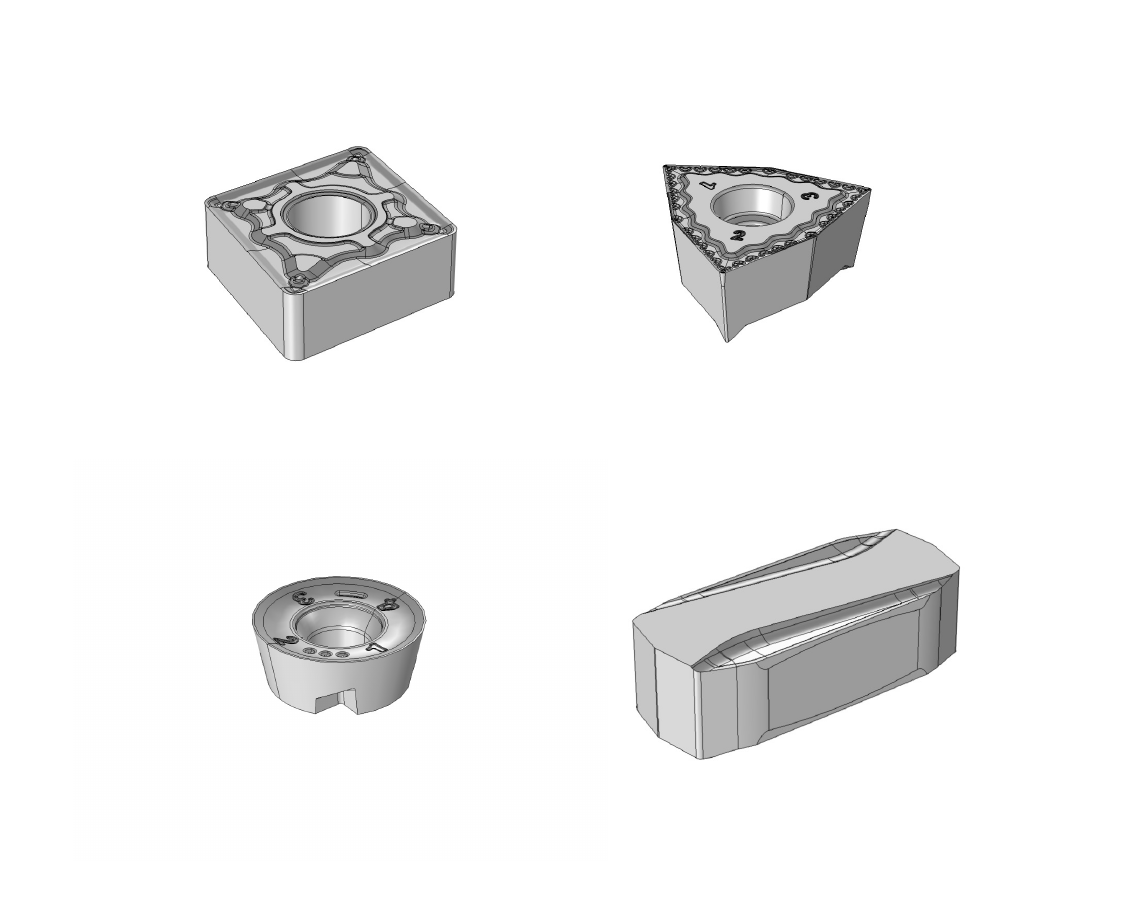}
   \caption{}
   \label{fig:inserts} 
\end{subfigure}
\hfill
\begin{subfigure}[b]{0.4\textwidth}
   \includegraphics[width=1\textwidth]{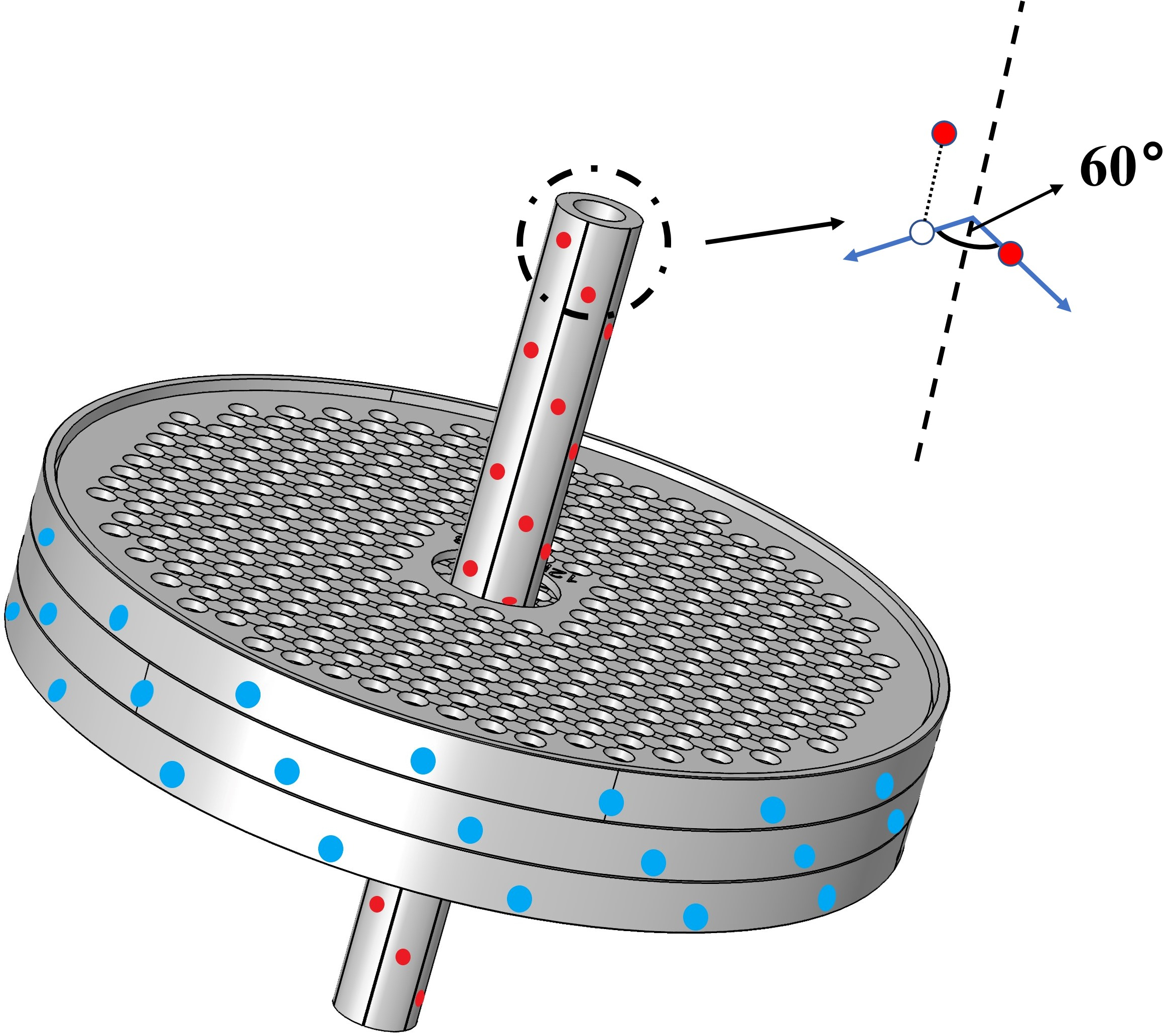}
   \caption{}
   \label{fig:3d-3disks}
\end{subfigure}
\caption{(a) Indicative geometries of the coated cutting tools. (b) A 3D representation of a 3-disk part of the reactor. The inlet perforations on the rotating inlet tube are shown in red. The outlet perforations for each disk are shown in blue.}
\label{fig:geometry-explain}
\end{figure}

The present work uses the example of a two-step coating process, which takes place in a commercial Chemical Vapor Deposition (CVD) reactor (Sucotec SCT600TH). The process is described in detail in previous works, which the interested reader can refer to \cite{papavasileiou2024integrating, papavasileiou2022efficient, papavasileiouEquationbasedDatadrivenModeling2023} and is summarized here for completeness.
First, a 9 $\mu$m Ti(C,N) layer is deposited on the cemented carbide cutting inserts, examples of which are presented in \cref{fig:inserts}. This step is followed by the deposition of an alumina layer under a AlCl\textsubscript{3}–CO\textsubscript{2}–HCl–H\textsubscript{2}–H\textsubscript{2}S chemical system, with the process temperature and pressure for the \alumina{} deposition being $T$=1005\textdegree{}C and $p$=80 mbar, respectively \citep{hochauerCarbonDopedAAl2O32012}.

The CVD reactor consists of 40-50 perforated disks stacked vertically, each accommodating inserts. \cref{fig:3d-3disks} provides a schematic 3D illustration of three such disks for clarity. Gas reactants enter the reactor through perforations in a rotating cylindrical tube positioned at the center of the disk stack. Each disk level features two diametrically opposite perforations, with a 60\textdegree{} angular offset between the axes connecting the inlet holes at each level. The rotating motion of the centrally located inlet tube, at a speed of 2 RPM, makes the process inherently periodic. An important aspect of the process is the fact that the internal geometry of the reactor varies between production runs as the geometry of the inserts and the disks on which they are placed is modified according to production requirements.

The primary objective of the process is to achieve a uniform coating thickness, as this uniformity directly correlates with consistent product longevity \citep{bar-henExperimentalStudyEffect2017,koronaki2014non}. Ideally, this uniformity would be consistent across all production runs, reactors, and sites. However, this is not always achieved. Therefore, it is crucial to develop a method for predicting the coating thickness of the inserts based on the reactor setup. In addition, establishing a systematic approach to evaluate factors affecting the uniformity of the coating thickness is essential. To this extent, the application of both equation-based methods \citep{papavasileiou2022efficient} and data-driven methods \citep{papavasileiouEquationbasedDatadrivenModeling2023, papavasileiou2024integrating} has been demonstrated in previous work, to which the interested reader is referred for further information on the process.

\section{The original data-driven approach}
\label{sec:original-approach}
\begin{figure}[!ht]
    \centering
    \includegraphics[width=.85\textwidth]{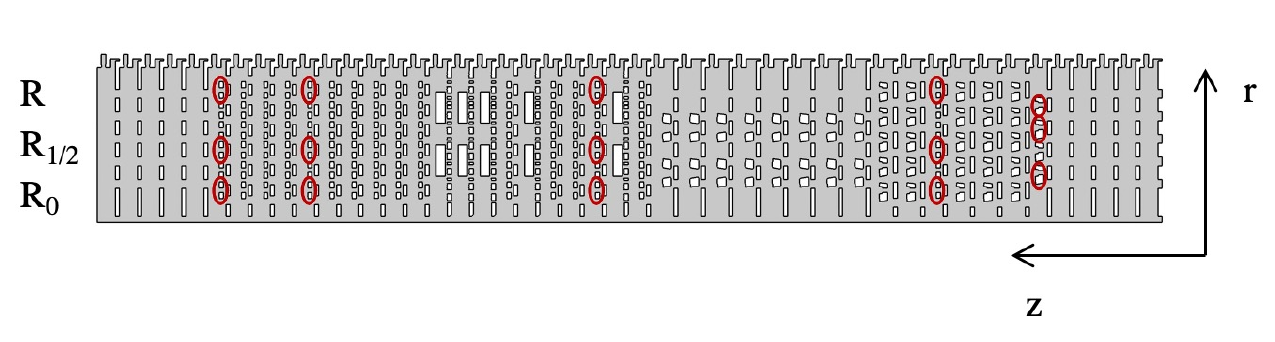}
    \caption{Positions with available \alumina{} thickness values from the production data for our test case. In general, across different production runs, the R position (the one closest to the reactor outlet) is the one with the highest amount of data. For this reason, the ML models are trained to make predictions for inserts placed in this position.}
    \label{fig:measurement-positions}
\end{figure}

At each production run, 15 thickness measurements are taken ex-situ, using the Calotest method \citep{lepickaInitialEvaluationPerformance2019}.
A 2D representation of the reactor indicating the points where thickness is typically measured is shown in \cref{fig:measurement-positions}. It should be noted though that, due to production decisions, additional measurements are taken in the R position (the one closest to the reactor outlet). For this reason, the focus of the \alumina{} coating thickness predictions falls on the R position of the reactor.

In addition to coating thickness measurements, the dataset also contains several features regarding the set-up of the production run. These features are used as inputs for our predictive ML models. An important feature is the production \enquote{recipe}, which encapsulates the steps taken and the process conditions during production. These specific details cannot be detailed here. Furthermore, there are features that contain information regarding the \enquote{how} and the \enquote{where} the inserts are placed within the reactor. More specifically, in the original approach, the following features are used: a) The number of inserts placed on each disk. b) The position of each disk within the reactor. c) The type of insert placed on each disk. Each type of insert has different geometrical characteristics. d) The surface area of the inserts placed on each disk.

\begin{figure}[!ht]
	\centering
	\includegraphics[width=0.75\linewidth]{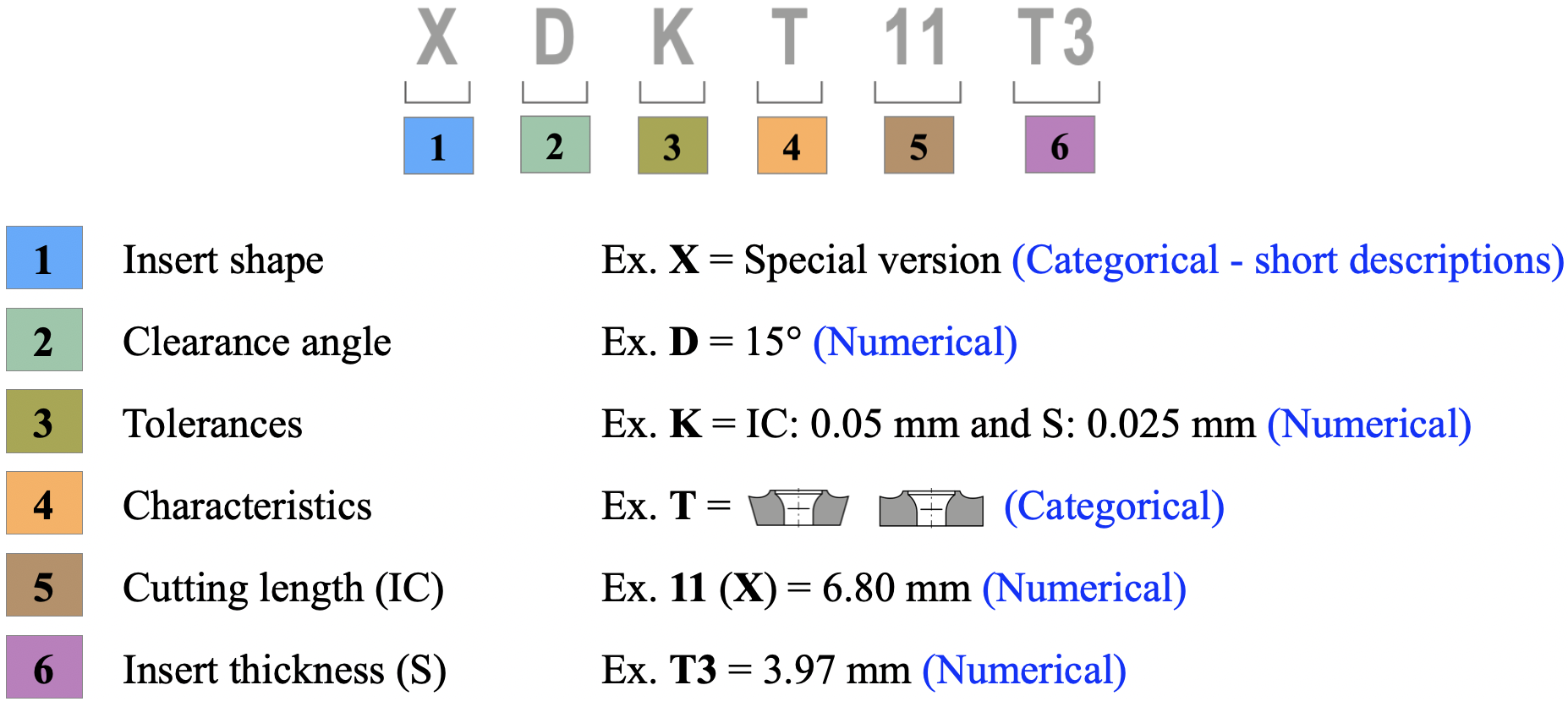}
	\caption{Example of an ISO designation for indexable inserts.}
	\label{fig:iso}
\end{figure}

It must be noted here that the categorical feature ``Insert geometry" corresponds to ISO designations for indexable inserts \cite{ecatalog}, which are codes composed of eight alphanumeric characters. Those characters represent some measurements (numerical features) or even textual descriptions (categorical features) related to the insert geometry, as shown in \cref{fig:iso}.

Additional features are engineered which include the total surface area and the standard deviation of the surface area of the to-be coated inserts. The information available for the neighboring disks, i.e. the disks above and below the disk of interest, are also used for the development of our predictive models. Subject matter expertise suggests that the difference between the nominal surface area indicated in the production \enquote{recipe} and the actual surface area of the inserts within the reactor, is an important feature.

\begin{table}[ht]
\centering
\caption{Summary of the features included in the training of the predictive models.}
\begin{tabular}{ccc}
\hline
\textbf{Feature}                                                                           & \textbf{Type}       & \textbf{Pre-processing} \\
\hline
Number of inserts on disk                                                                  & Numerical (integer)& standardization        \\
Surface area of inserts on disk                                                            & Numerical (float)   & standardization        \\
Disk position                                                                           & Numerical (integer)  & standardization        \\
\begin{tabular}[c]{@{}c@{}}Total surface area of inserts\\ inside the reactor\end{tabular} & Numerical (float)& standardization        \\
Surface area standard deviation                                                            & Numerical (float)   &  standardization        \\
\begin{tabular}[c]{@{}c@{}}Nominal “recipe” surface area \\ - actual surface area\end{tabular}    & Numerical (float)   &  standardization        \\
Production “recipe”                                                                        & Categorical      & binary encoding        \\
\textcolor{gray}{Insert geometry}                                                                            & Categorical     & binary encoding        \\
\textcolor{gray}{Insert geometry – disk above}                                                               & Categorical      & binary encoding        \\
\textcolor{gray}{Insert geometry – disk below}                                                               & Categorical      & binary encoding        \\
\hline
\end{tabular}
\label{table:feature-summary}
\end{table}

Following this step, ten features, both numerical and categorical, are available for the development of the predictive model, as presented in \cref{table:feature-summary}. The numerical features are standardized: centered (subtraction of the mean) and scaled (divided by the standard deviation), while the categorical variables are encoded using binary encoding \citep{potdarComparativeStudyCategorical2017}. 

The XGBOOST model (cf. \Cref{xgb}) is the most efficient and high-performing ML model for this task, as implemented and shown in \cite{papavasileiou2024integrating}. Therefore, this model will also be considered for the present work.

\section{Methods}\label{Methods}

This section provides a succinct description of the methods and algorithms used to create vector representations of insert shape descriptions through various embeddings, and to present similarities between the different representations. These vectors are included as numerical features in a predictive model that predicts insert thickness based on other insert features listed in \cref{sec:original-approach}. Additionally, the model performance will be evaluated and compared across different embeddings using a cross-validation method, followed by feature importance and Shapley analysis.

\subsection{Embeddings and transformers}\label{Embtrans}

Embedding is a method used to convert words or sentences from a vocabulary into dense numerical vectors suitable for machine learning models \cite{doi:10.1080/01638539809545028, mikolov2013efficientestimationwordrepresentations}. More precisely, each word or sentence is mapped to a vector in Euclidean space, known as the embedded latent space. This representation helps capture and understand the semantic similarity between words.

In addition, transformers are a type of neural network architecture \cite{NIPS20173f5ee243} that include an embedding layer, which encodes semantically similar words or sentences as close elements in the latent space \cite{dar2023analyzingtransformersembeddingspace}. Then, these deep learning methods rely on attention mechanisms to process sequential data. In particular, transformers are designed to manage context more effectively when processing text, allowing each word in a sentence to connect with every other word. This ability is essential for understanding meaning within context.

That said, we will now focus on the models used here: \textit{Doc2Vec} and two pre-trained models (\textit{all-MiniLM-L12-v2} and \textit{all-mpnet-base-v2}), respectively, whose main characteristics are summarized in \cref{SumLLM}.

\vspace{-0.5 cm}
\begin{table}[ht]
	\begin{center}
		\caption{Summary of the NLPs}
		\begin{adjustbox}{width=0.825\textwidth}
		\begin{tabular}{ccccc}
			\hline
			\textbf{Model} & \textbf{Dimension} & \textbf{Pre-trained Model} & \textbf{Training Data} \\
			\hline
			Doc2Vec & set to 3 & no & available insert shape descriptions \\
			all-MiniLM-L12-v2 & 384 & yes & 1B+ training pairs \\
			all-mpnet-base-v2 & 768 & yes  & 1B+ training pairs \\
			\hline
		\end{tabular}
	\end{adjustbox}\label{SumLLM}
	\end{center}
\end{table}

Both transformers were taken from Hugging Face Hub, which contains a vast collection of public Sentence Transformers models, extensively evaluated for their ability to embed sentences. The \textit{all-mpnet-base-v2} model provides the best quality and embedding performance, while \textit{all-MiniLM-L12-v2} is three times faster and still offers good embedding quality. It is also the best-performing model among those with a smaller latent space.

Existing studies evaluated these models on various benchmark problems \cite{lau2016empiricalevaluationdoc2vecpractical, devlin2019bertpretrainingdeepbidirectional, song2020mpnetmaskedpermutedpretraining, wang2020minilmdeepselfattentiondistillation, pretrained}, consistently demonstrating strong performance, especially in sentence similarity tasks. The results and selected pre-trained models are available in the Hugging Face open-access transformers repository, where the top two models were chosen based on performance scores \cite{huggingfaces} 

Non-pretrained models like \textit{Doc2Vec} \cite{mahajan2023dauntingdilemmasentenceencoders} generate similar vector representations for related categories, enhancing categorical variable integration in machine learning, especially in industrial settings where these variables reflect key interdependencies \cite{8718215} (e.g., material types, operating modes, tool geometries). \textit{Doc2Vec} also allows control over latent space dimensionality, balancing efficiency and performance. Meanwhile, pre-trained models offer robust sentence embeddings that capture contextual relationships. Chosen for their embedding quality, efficiency, and scalability, these models effectively categorize and cluster industrial categorical variables \cite{WU2006220, Miao03042023}.

\subsubsection{Doc2Vec}

\textit{Doc2Vec} \cite{ref:Doc2Vec} is an unsupervised machine learning algorithm from the \textit{Gensim library} to obtain vector representations of documents. In this case, a document is an object of the text sequence type and it could be anything from a short character string to documents structured by paragraphs such as an article or a book. 

This algorithm is an extension of \textit{Word2Vec}, based on the \textit{bag-of-words} \cite{mikolov2013efficientestimationwordrepresentations} and \textit{bag-of-n-grams} \cite{mikolov2013efficientestimationwordrepresentations, mikolov2013distributedrepresentationswordsphrases}. It was designed to create fixed-length vector representations from pieces of texts. This model allows us to capture semantic meanings (distances between the words) and context (relationships between words and documents), which increase the accuracy of various tasks such as document classification, clustering, and recommendation.

Initially, we pre-process the documents to ensure agreement with the \textit{Doc2Vec} model. This preprocessing includes the removal of punctuation and special characters, tokenization (word by word), and the creation of a dictionary. These steps are crucial to standardize the text and reduce noise, thereby improving the quality of the data and the resulting vectors.

The training process involves some parameters with which we can experiment to evaluate the model performance.

\begin{itemize}
	\item \texttt{vector\_size}: the dimensionality of the feature vectors.
	\item \texttt{min\_count}: words that appear less than $n \in \mathbb{N}$ times are excluded from the training to focus on more frequent terms.
	\item \texttt{epochs}: the model is trained for $N$ epochs to ensure adequate learning of patterns in the text.
\end{itemize}

\subsubsection{all-MiniLM-L12-v2}

The \textit{all-MiniLM-L12-v2} \cite{wang2020minilmdeepselfattentiondistillation, huggingfaces} is a compact and efficient pre-trained language model designed for a variety of  natural language processing (NLP) tasks \cite{BERT}. It has been fine-tuned with more than 1 billion textual pairs to balance computational efficiency with high performance \cite{devlin2019bertpretrainingdeepbidirectional}.

In addition, it utilizes a 12-layer deep network with attention mechanisms that allow to capture contextual relationships within text sequences efficiently. This understanding is essential for text classification, sentiment analysis, and named entity recognition tasks.

Pre-processing techniques also involve tokenization, which breaks down input text into meaningful units for the model to process, and data augmentation strategies to increase the model's ability to generalize \cite{pretrained}. Then, the model produces embeddings of fixed length equal to 384, which are included in the training of our predictive model.

\subsubsection{all-mpnet-base-v2}

The \textit{all-mpnet-base-v2} model \cite{song2020mpnetmaskedpermutedpretraining, huggingfaces} is a pre-trained transformer-based architecture developed by Microsoft. The \textit{all-mpnet-base-v2} model operates with a transformer encoder comprising 12 layers, each with 12 attention heads, and a hidden size of 768. It has been fine-tuned with more than 1 billion textual pairs to capture a wide array of linguistic patterns and contextual representations, making it highly suitable for various NLP tasks \cite{BERT, devlin2019bertpretrainingdeepbidirectional}.

Pre-processing involves tokenizing the input text, which is then fed into the model to produce embeddings of length 768. These embeddings are utilized in our experiments to explore various aspects of language understanding and contextual relevance.

\subsubsection{Similarity between descriptions}\label{Sec:Similarity}

Cosine similarity is a widely used metric in NLP for assessing the similarity between two vectors. In the context of sentences embeddings, cosine similarity is a valuable method for comparing the similarity of words or documents \cite{Steck2024, scikitlearn}.

It measures the cosine of the angle between two vectors in a high-dimensional space. The similarity score ranges from -1 to 1, where a score of -1 signifies completely opposite vectors, 0 indicates no similarity, and 1 means identical vectors. The cosine similarity (\textit{Sim}) between two vectors is computed using the following formula:
$$\textit{Sim} (A,B) = \dfrac{A \cdot B}{\|A\| \|B\|},$$

where $A \cdot B$ represents the dot product between vectors $A$ and $B$, while $\|\cdot\|$ denotes the Euclidean norm. 
In Fig. 4, the calculation of cosine similarity is illustrated, along with the pairwise computation of the similarity scores derived from the vector representations generated by embeddings and transformers.

\begin{figure}[!ht]
	\centering
	\includegraphics[width=1\linewidth]{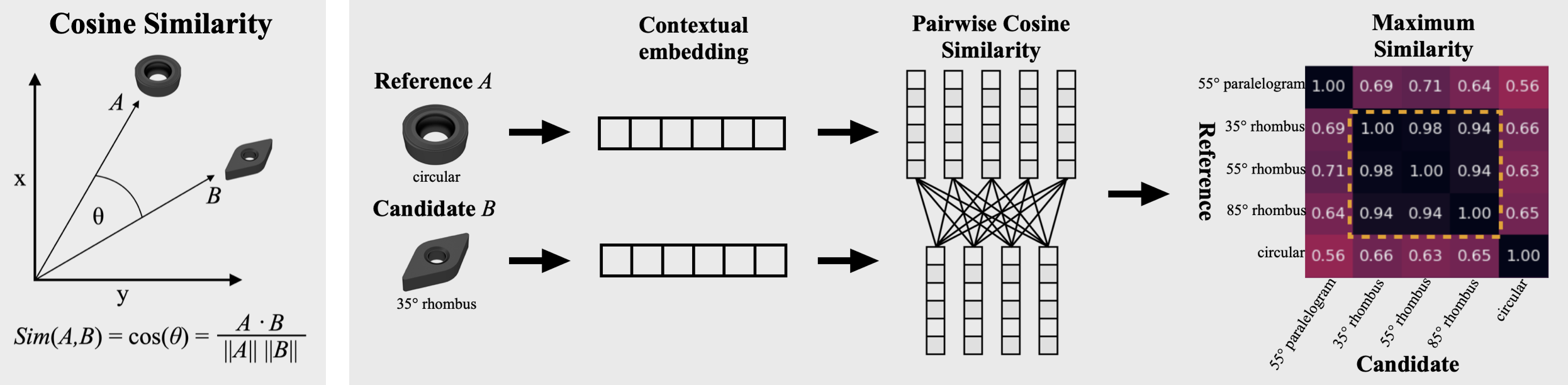}
	\caption{Cosine similarity is defined as the cosine of the angle between the embedding vectors of two objects. Using this concept, we consider two insert shapes (reference and candidate) that are embedded into a feature space, where their pairwise cosine similarity is computed. The similarity matrix on the right contains all the computed values and illustrates the relationships between different insert shape embeddings.}
	\label{fig:cossim}
\end{figure}

\subsection{Dimensionality reduction}

To effectively handle the high dimensionality of embedding spaces generated by different transformers (cf. Sections \ref{Embtrans} and \ref{results}), we employ linear dimensionality reduction methods, such as \textcolor{blue}{Principal Component Analysis (PCA)} \cite{HOTELLING1933, MACKIEWICZ1993}, which is based on singular value decomposition. This technique reduces the original data while preserving its overall variability. Such an approach has been successfully applied for dimensionality reduction in the context of CVD processes \cite{ISAAC14, KORONAKI2019}.
Alternatively, nonlinear methods like Uniform Manifold Approximation and Projection (UMAP) \cite{mcinnes2020umapuniformmanifoldapproximation, ghojogh2021uniformmanifoldapproximationprojection}, a manifold learning technique for dimensionality reduction, can be employed. UMAP is also widely used in chemical engineering for data visualization \cite{ROVIRA2022135250,JOSWIAK2019104189}.

\subsection{XGBOOST}\label{xgb}
Following previous work \cite{papavasileiouEquationbasedDatadrivenModeling2023}, where an XGBOOST (Extreme Gradient Boosting) regressor \citep{chenXGBoostScalableTree2016} demonstrated excellent performance, XGBOOST is also used in the present implementation. XGBOOST is an ensemble \citep{hastieEnsembleLearning2009} tree-based method \cite{jamesTreeBasedMethods2021} that allows both classification and regression. The XGBOOST algorithm builds shallow trees sequentially, with each tree being fit using the error residuals (difference between actual and predicted values) of the previous model. For more information on XGBOOST, the reader is referred to the original paper by \citet{chenXGBoostScalableTree2016}.

\subsubsection{K-fold cross-validation}

To ensure reliable predictions and avoid overfitting, it is common practice to reserve a portion of the data as a test set. However, if the test set is used for other purposes, such as fine-tuning hyperparameters, a validation set is also needed: the model is trained on the training set, validated on the validation set, and finally evaluated on the test set \cite{scikitlearn}. 

However, splitting the data into three sets can reduce the amount of data available for training. The $k$-fold cross-validation \cite{K-fold} manages this by dividing the data into $k$-folds. The model is trained on $k-1$ folds and tested on the remaining fold. This process is repeated $k$ times, and the final performance measure is the average across all folds.

Therefore, this approach is more efficient in treating the available data and addressing the randomness of the test-set selection.

\subsubsection{Feature Importance}

Feature importance is a method used to measure the significance of input features in predicting a target variable by assigning them a score based on their impact \cite{theFI}. Various techniques can be employed to determine feature importance, including statistical correlation measures \cite{10.1093/bioinformatics/btq134}, coefficients from linear models \cite{doi10.1198tast.2009.08199}, and decision tree-based methods \cite{doi10.1198tast.2009.08199}.

In this study, we utilize the XGBOOST model to analyze feature importance \cite{XGBFI}. Thus, after constructing the boosted trees, it provides a way to retrieve the importance scores for each feature. 
In particular, we will use the \texttt{total\_gain} score, which represents the total contribution of a feature to the model, calculated by considering the contribution of each feature across all trees \cite{UGUZ20111024, gainFI2}. A higher \texttt{total\_gain} value compared to other features indicates that the feature plays a more significant role in generating predictions.

\subsubsection{SHAP Analysis}
Shapley values, introduced by \citet{shapleyValueNPersonGames1952} and later applied to machine learning models in \citep{lundbergUnifiedApproachInterpreting2017, lundbergLocalExplanationsGlobal2020}, assess the average contribution of each feature to predictions. This helps us to understand how changes to a variable affect the model's output. The core concept behind Shapley value-based explanations in machine learning is to allocate credit for a model’s output fairly among its input features, based on cooperative game theory principles. In this framework, the input features are treated as players in a game, where each player can either participate or not. If a player (input feature) joins, its value is known; if not, its value remains unknown. Shapley values are additive, meaning that in the context of model explanation, the sum of all SHAP values across the input features will always equal the difference between the baseline (expected) output and the actual output for the prediction being explained.

In this work, the SHAP (SHapley Additive exPlanations) analysis is applied to the proposed XGBOOST regression models. The resulting Shapley values will highlight the importance and influence of each feature on the model output.

Finally, to obtain an aggregated average score of the Shapley values, which allows us to derive the average contributions of each feature, we first sum the rows corresponding to each feature. Next, we take the absolute value of the resulting vector to ensure a positive vector of individual contributions. Then, we compute the average of these contributions.

\section{Results}\label{results}


\subsection{Data preprocessing}


Regarding the insert names, each ISO designation code (cf. \cref{fig:iso}) is broken down into its various features, aiming to provide additional information that could be useful to include in the XGBOOST model. Additionally, it is important to note that the other categorical feature, production \enquote{recipe}, is independent of the inset shape. This is because the reactor contains various inserts with different shapes, making this feature irrelevant when employing embeddings.

\begin{table}
\centering
	\caption{Summary of features included in the training of the predictive models after pre-processing the insert geometries.}
	\begin{adjustbox}{width=0.96\textwidth}
		\begin{tabular}{ccc}
			\hline
			\textbf{Feature}                                                                           & \textbf{Type}       & \textbf{Pre-processing} \\
			\hline
			Number of inserts on disk                                                                  & Numerical (integer)& standardization        \\
			Surface area of inserts on disk                                                            & Numerical (float)   & standardization        \\
			Disk position                                                                           & Numerical (integer)  & standardization        \\
			\begin{tabular}[c]{@{}c@{}}Total surface area of inserts\\ inside the reactor\end{tabular} & Numerical (float)& standardization        \\
			Surface area standard deviation                                                            & Numerical (float)   &  standardization        \\
			\begin{tabular}[c]{@{}c@{}}Nominal “recipe” surface area \\  – actual surface area\end{tabular}    & Numerical (float)   &  standardization        \\
			\textcolor{Emerald}{\texttt{Insert clearance angle}}                                                                            & Numerical (float)    & standardization        \\
            \textcolor{Emerald}{\texttt{Insert clearance angle – disk above}}                                                                            & Numerical (float)    & standardization        \\
            \textcolor{Emerald}{\texttt{Insert clearance angle – disk below}}                                                                            & Numerical (float)    & standardization        \\
			\textcolor{RawSienna}{\texttt{Insert cutting length}}                                                                           & Numerical (float)    & standardization        \\
			\textcolor{RawSienna}{\texttt{Insert cutting length – disk above}}                                                                           & Numerical (float)    & standardization        \\
            \textcolor{RawSienna}{\texttt{Insert cutting length – disk below}}                                                                           & Numerical (float)    & standardization        \\
            \textcolor{Purple}{\texttt{Insert thickness}}                                                                            & Numerical (float)    & standardization        \\
			\textcolor{Purple}{\texttt{Insert thickness – disk above}}                                                                            & Numerical (float)    & standardization        \\
			\textcolor{Purple}{\texttt{Insert thickness – disk below}}                                                                            & Numerical (float)    & standardization        \\
            \textcolor{olive}{\texttt{Insert cutting length tolerance}}                                                                           & Numerical (float)    & standardization        \\
			\textcolor{olive}{\texttt{Insert cutting length tolerance – disk above}}                                                                           & Numerical (float)    & standardization        \\
            \textcolor{olive}{\texttt{Insert cutting length tolerance – disk below}}                                                                           & Numerical (float)    & standardization        \\
            \textcolor{olive}{\texttt{Insert thickness tolerance}}                                                                            & Numerical (float)    & standardization        \\
			\textcolor{olive}{\texttt{Insert thickness tolerance – disk above}}                                                                            & Numerical (float)    & standardization        \\
			\textcolor{olive}{\texttt{Insert thickness tolerance – disk below}}                                                                            & Numerical (float)    & standardization        \\
			Production “recipe”                                                                        & Categorical      & binary encoding        \\
			\textcolor{Cerulean}{\texttt{Insert shape}}                                                                            & Categorical    & Embedding* + standardization        \\%
			\textcolor{Cerulean}{\texttt{Insert shape – disk above}}                                                               & Categorical      & Embedding* + standardization        \\
			\textcolor{Cerulean}{\texttt{Insert shape – disk below}}                                                               & Categorical      & Embedding* + standardization        \\
			\textcolor{YellowOrange}{\texttt{Insert characteristic}}                                                                            & Categorical    & binary encoding        \\
			\textcolor{YellowOrange}{\texttt{Insert characteristic – disk above}}                                                                            & Categorical    & binary encoding        \\
			\textcolor{YellowOrange}{\texttt{Insert characteristic – disk below}}                                                                            & Categorical    & binary encoding        \\
			\hline
		\end{tabular}
		\label{table:feature-summaryD2V}
	\end{adjustbox}
	\caption*{{\footnotesize *\textit{Doc2Vec}, \textit{all-mpnet-base-v2} and \textit{all-MiniLM-L12-v2}.}}
 \end{table}

The new features  are summarized in \cref{table:feature-summaryD2V}. In the data transformation stage, we first apply data imputation to ensure that missing values were not included in our analysis. Then, we standardize the numerical features and use embeddings/transformers to represent categorical features as vectors.

Subsequently, the \textit{Doc2Vec} model is trained for 2000 epochs. The vocabulary used for training consists of at least twelve short descriptions of insert shapes distributed across the current disk position and the positions above and below it. Therefore, we set the dimension of the dense vectors to the floor of the average number of words used in each description: \texttt{vector\_size} = 3, and set the \texttt{min\_count} parameter to 1 due to the short length of these descriptions.

In contrast, for the two pre-trained transformers, we use their most recent versions, which were trained on over one billion text pairs. Moreover, the resulting dense vectors have fixed and predetermined sizes: 384 for the \textit{all-MiniLM-L12-v2} model and 768 for the \textit{all-mpnet-base-v2} model.

\subsection{Similarity}\label{sec:Similarity}

The first step is to verify whether the vectors that represent the embeddings by the different models are truly capturing the meaning of the sentences, including both semantic and contextual. Recall that each sentence has been mapped as vectors in Euclidean space. Consequently, we can consider measures such as the distance between these vectors or the angle between two vectors, as indicators of sentence similarity (see \cref{Sec:Similarity}).

\begin{figure}[!ht]
	\centering
	\includegraphics[width=0.7\linewidth]{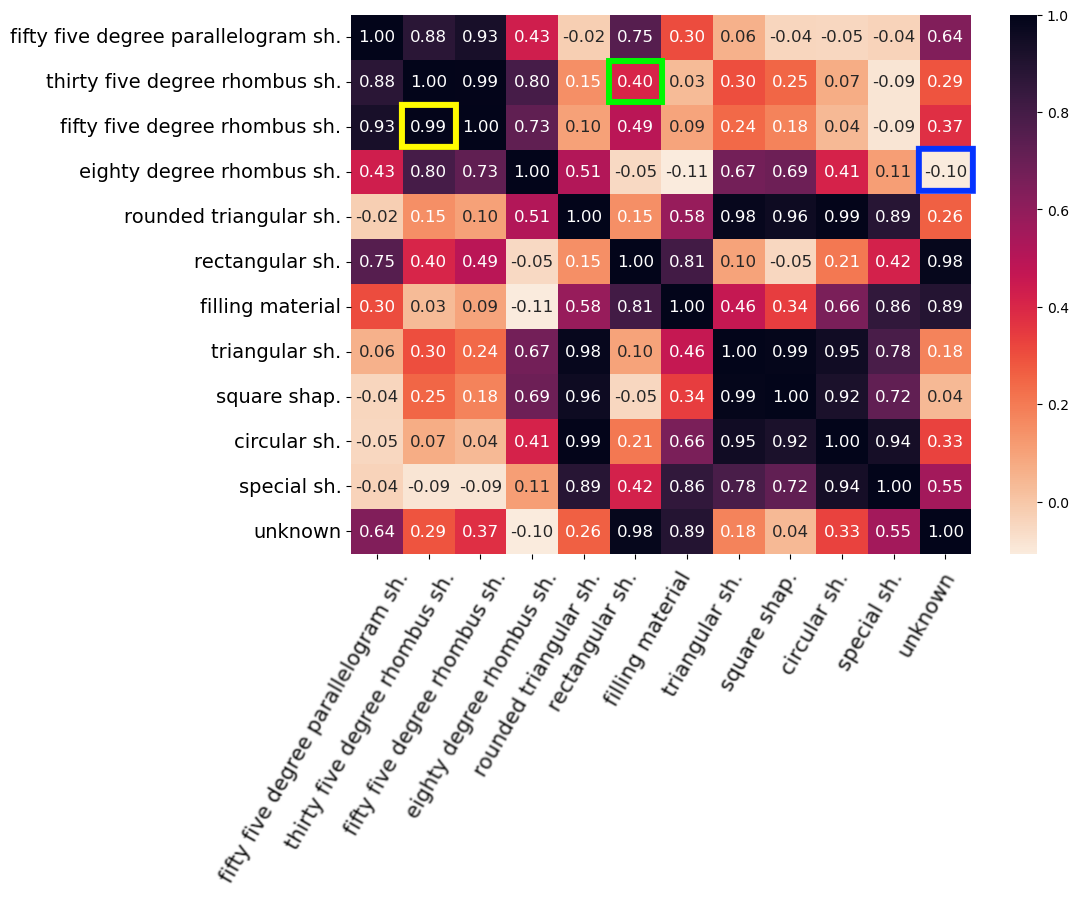}
	\caption{Heatmap showing the cosine similarity values calculated from the dense vectors obtained using the \textit{Doc2Vec} model. These values range from -1 to 1, where values closer to 0 are represented in lighter shades, and values closer to -1 or 1 are colored using darker shades. The value enclosed in a yellow box represents the high similarity between rhombus-shaped objects; in contrast, the low values enclosed in the green and blue boxes suggest lower similarity between the respective shapes. }
	\label{fig:s1}
\end{figure}

In this section, we propose to study the values obtained by calculating the $\texttt{cosine\_similarity}$ between each pair of dense vectors associated with the insert shape descriptions implemented in this study. These values are then collected and presented as heatmaps for each embedding and transformer. 

As an example, let us consider the following cases. In \cref{fig:s1}, the value enclosed in the yellow square indicates that the fifty-five-degree rhombus has a high similarity in shape characteristics to the thirty-five-degree rhombus. In the case of intermediate values, such as the one enclosed in the green square, it signifies that the thirty-five-degree rhombus exhibits some common shape attributes with the rectangular insert. Finally, the lower value, enclosed in a blue square, confirms a significantly different shape description, illustrating the distinction between the unknown insert shape and an eighty-five-degree rhombus insert shape.

Concretely, for the \textit{Doc2Vec} model (see \cref{fig:s1}), we first observe a sort of reflexive property between descriptions, meaning each description is perfectly similar to itself. Additionally, we can see certain groups of descriptions that are expected to be similar, such as the group of rhombus-shaped inserts with different degrees. We also notice specific connections between certain shapes, like triangular and circular shapes with the rounded triangular shape, or some similarity among standard shapes like triangles, circles, and squares. It is also interesting to note that the \textsf{filling material} is close to the \textsf{unknown} inserts shape. In this context, the \textsf{filling material} refers to a non-specified insert shape that occupies the void in a disc, ensuring uniformity during the deposition process and maintaining the structural integrity of the disc. Therefore, these results illustrate that the \textit{Doc2Vec} model is capturing reasonably well the semantic and contextual meaning of the descriptions.

\begin{figure}[!ht]
	\centering
	\includegraphics[width=0.7\linewidth]{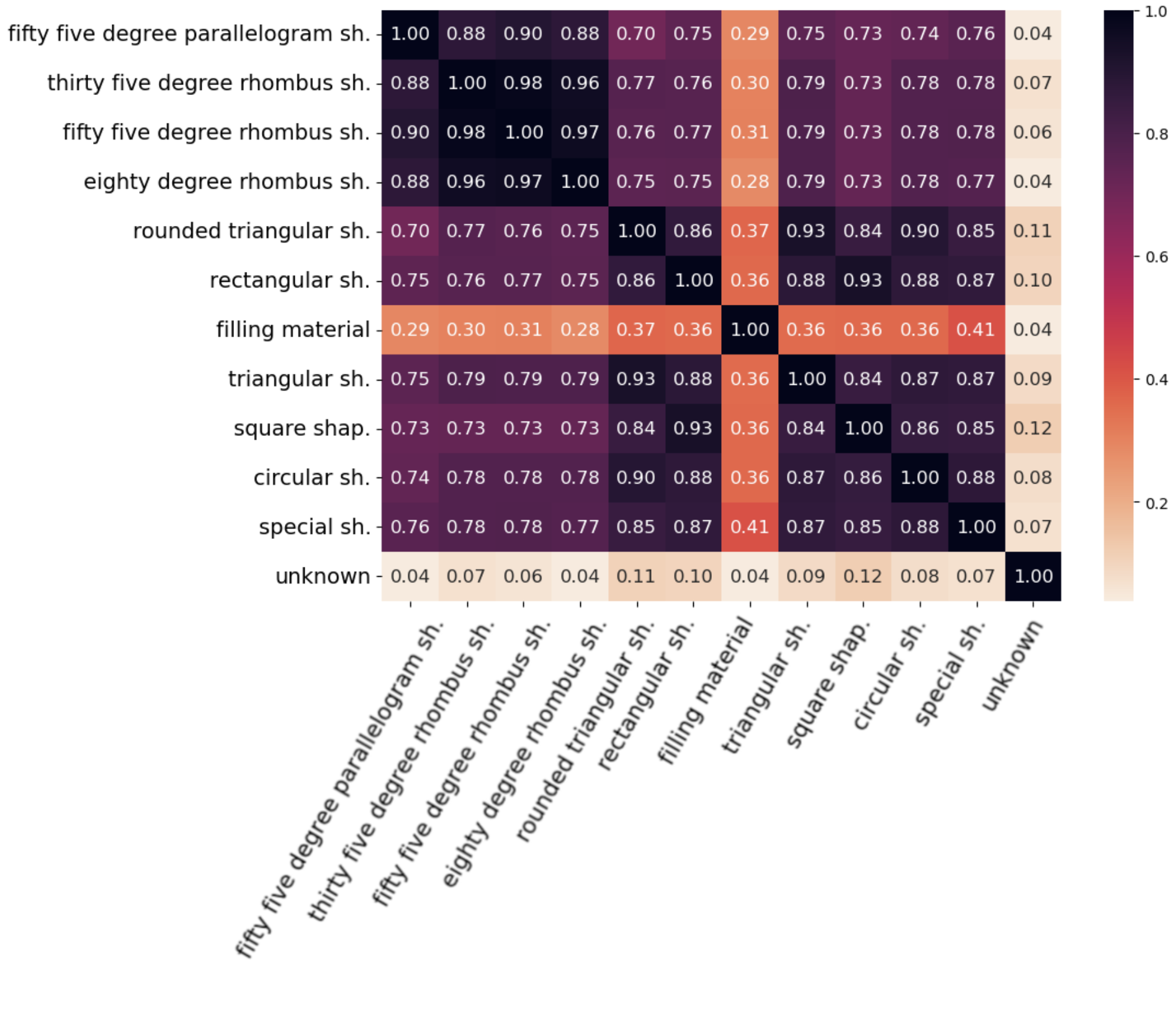}
	\includegraphics[width=0.7\linewidth]{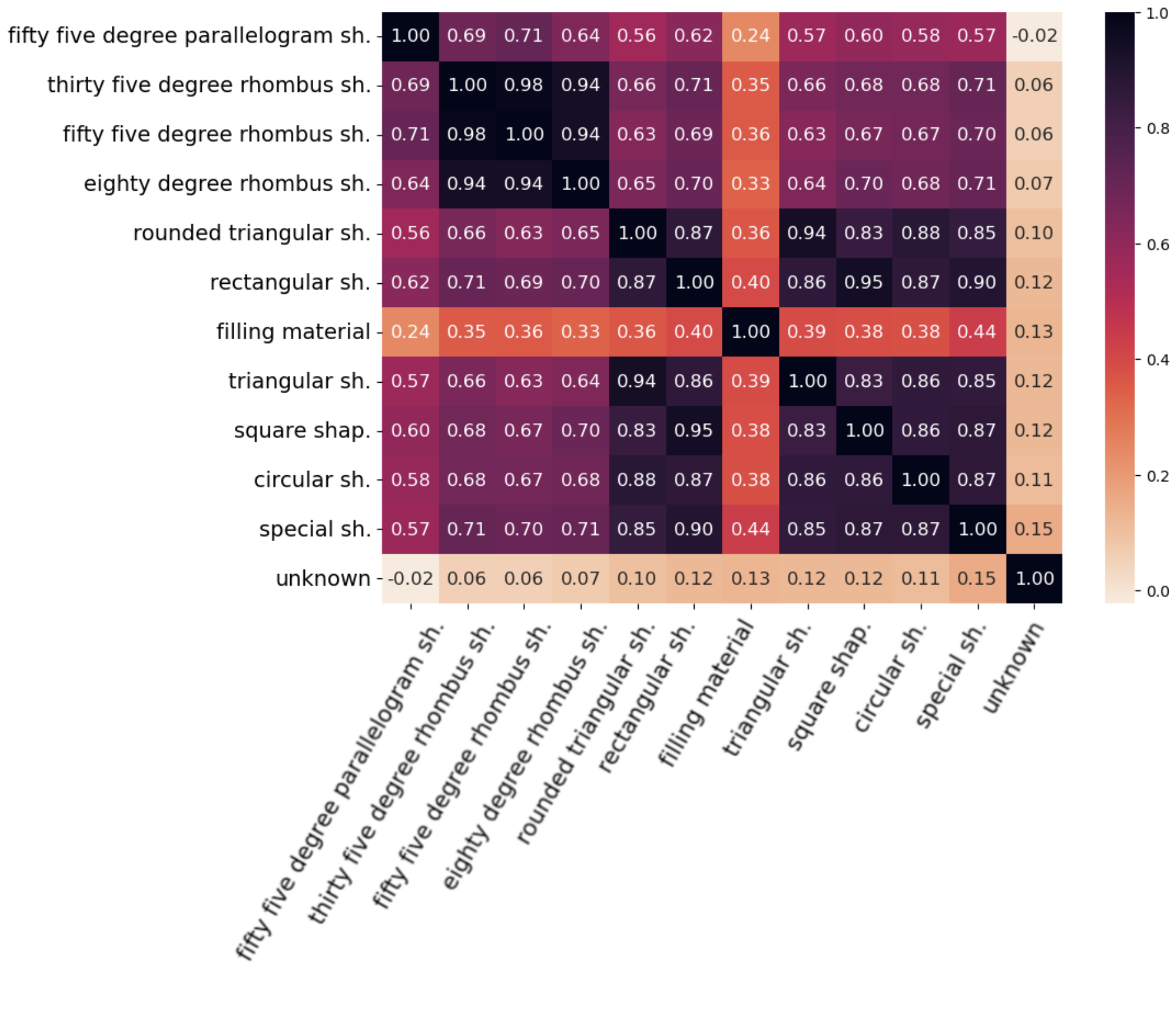}
	\caption{Heatmaps showing cosine similarity values computed for each of the embeddings generated by the \textit{all-MiniLM-L12-v2} model (top) and the \textit{all-mpnet-base-v2} model (bottom). The values range from -1 to 1, where lighter shades correspond to lower similarity and darker shades to higher similarity. Shortened versions of the twelve embedded descriptions are used as references on both the horizontal and vertical axes.}
	\label{fig:s2}
\end{figure}

Additionally, in the case of the \textit{all-MiniLM-L12-v2} model, we again observe the reflexivity property. We also see similar patterns to those of the previous case, but more marked (refer to \cref{fig:s2}). In the  with values surpassing 0.83, among semantically close insert shape descriptions, as expected. For instance, in \cref{fig:s2} (figure above) \texttt{cosine\_similarity} values for rhombus or triangular shapes. Two additional observations stand out: the corresponding similarity values for the descriptions \textsf{filling material} and \textsf{unknown shape} with respect to other descriptions are below 0.44 in the first case and closer to zero in the second case. This indicates that the model is correctly capturing their meanings, and we have a reasonably accurate representation of the considered descriptions.
\vspace{-0.2 cm}

\newpage
\begin{figure}[!ht]
	\centering
		\includegraphics[width=0.7\linewidth]{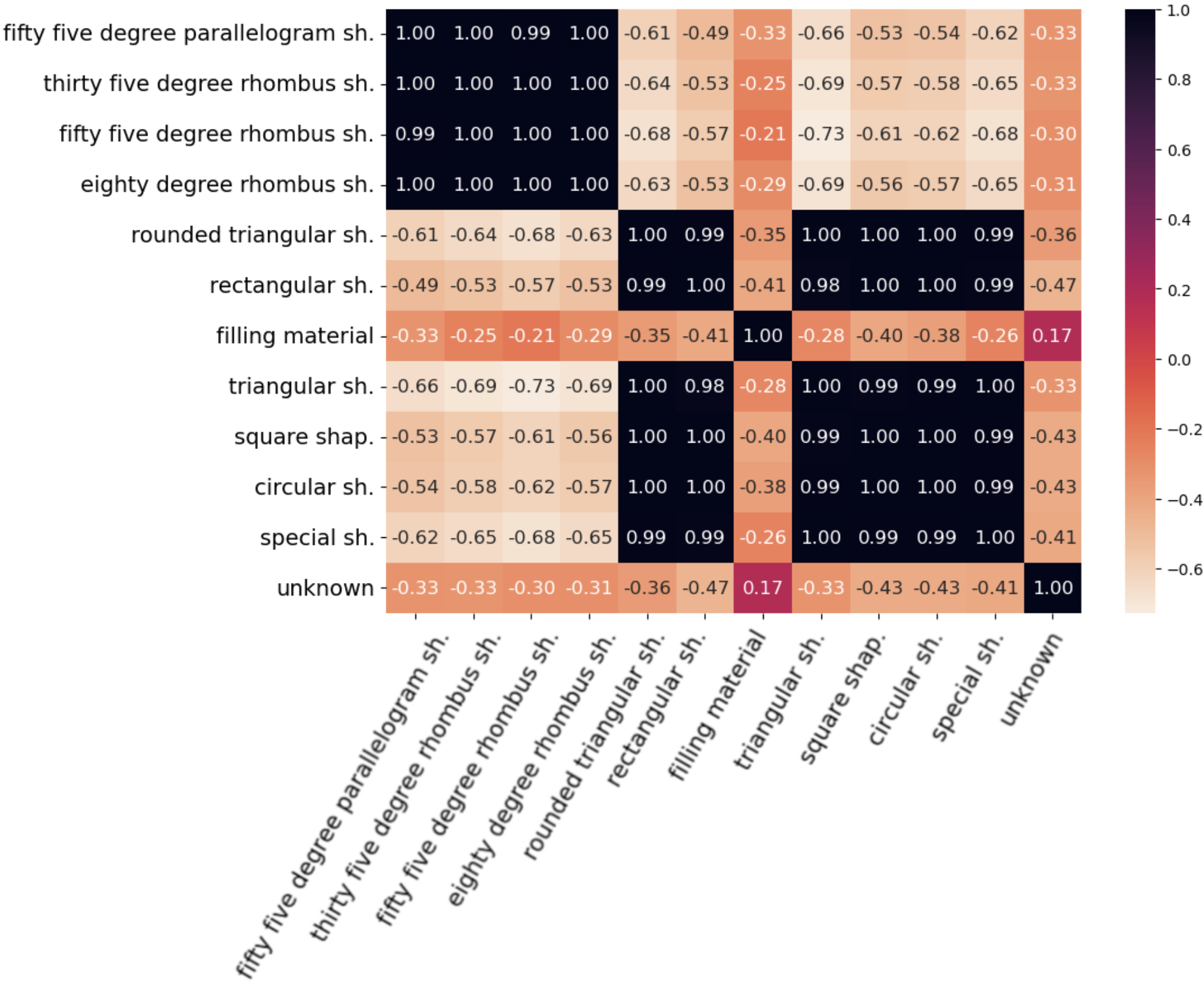}
		\includegraphics[width=0.7\linewidth]{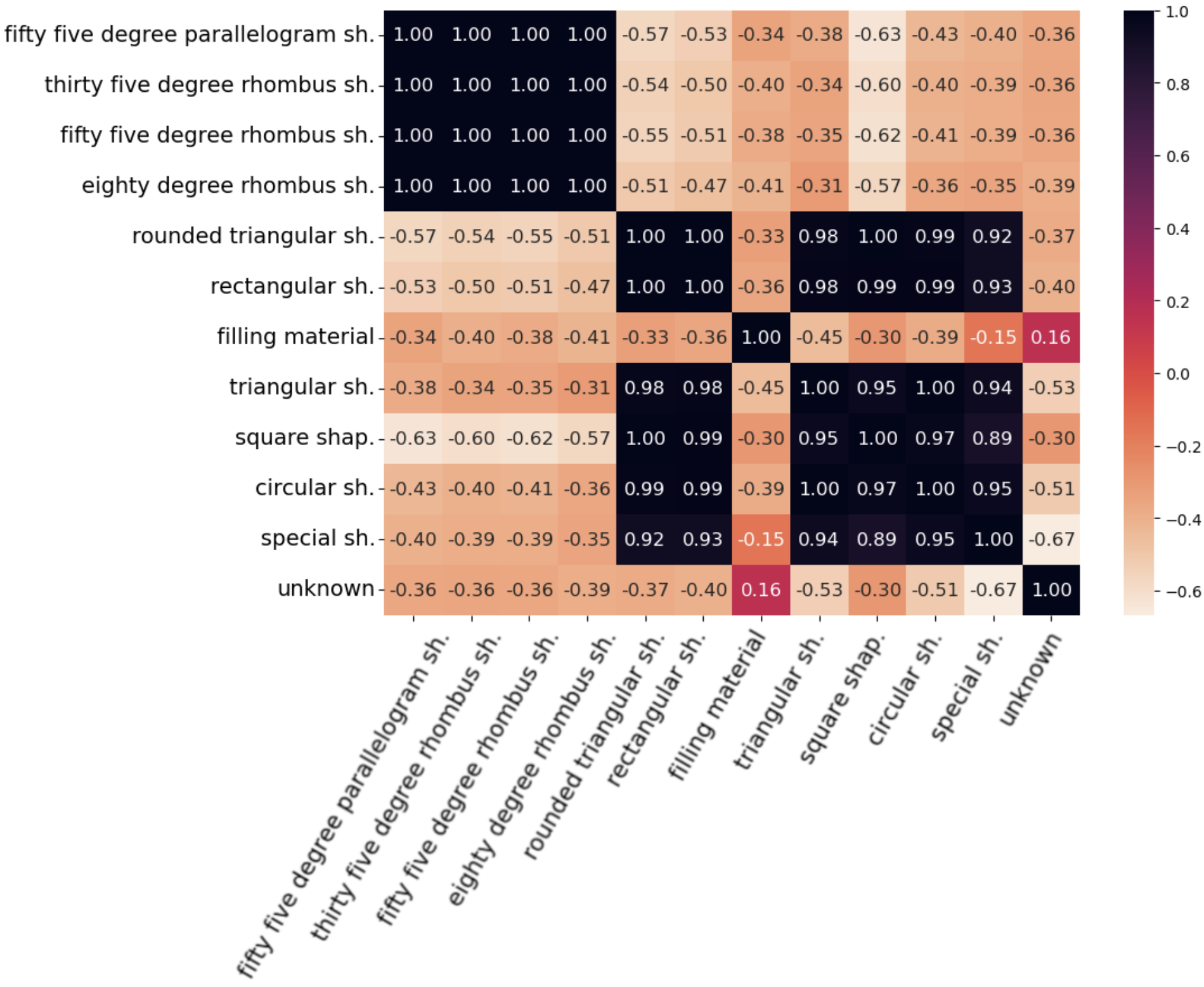}
	\caption{Heatmaps showing cosine similarity values computed for each of the dense vectors generated by the \textit{all-MiniLM-L12-v2} model (top) and the \textit{all-mpnet-base-v2} model (bottom), both after applying PCA. The values range from -1 to 1, where lighter shades correspond to lower similarity and darker shades to higher similarity. Shortened versions of the twelve embedded descriptions are used as references on both the horizontal and vertical axes.}
	\label{fig:s3}
\end{figure}

\newpage
For the \textit{all-mpnet-base-v2} model, we obtain results similar to those of the \textit{all-MiniLM-L12-v2} model, maintaining the same patterns (see \cref{fig:s2}). In \cref{fig:s2} (figure bellow), the values for descriptions such as \textsf{filling material} or \textsf{unknown shape} are distinct from other groups, with values below 0.41. Additionally, there is a high similarity among the descriptions in the group of quadrilaterals located in the upper left of \cref{fig:s2}, with values exceeding 0.88. However, there is no clear differentiation between standard shapes such as squares, circles, or triangles, which have values in the middle and lower right part of this \cref{fig:s2}. 

Now, it is worth noting that one of the advantages of the \textit{Doc2Vec} model is that, unlike pre-trained transformers, we can vary the dimensionality of the latent space according to our needs. We first compare the implementation of \textit{Doc2Vec} using latent spaces of larger size, specifically we select 384 and 768, i.e. the dimension of the embedding of the two pre-trained models, to the implementation of \textit{Doc2Vec} with a three-dimensional embedding.

Thus, two more possible scenarios are interesting for further study of \texttt{cosine\_similarity} at this point: the first regarding \textit{Doc2Vec}, for which we can consider latent spaces with dimensions of 384 and 768, as in the case of both transformers. The results obtained are similar to those for a fixed dimensionality of 3. 

In contrast, the second scenario corresponds to the high dimensionality of the latent spaces produced by both transformers. Possible consequences of reducing their dimensionality can be explored. For this study, we apply a linear dimensionality reduction method, PCA, to compress the data into three dimensions (based on an analysis of the reconstruction error), and a non-linear method, UMAP, with the same dimensionality.
Although UMAP is typically more efficient for dimensionality reduction, in this case, PCA provides a better representation by preserving the semantic and contextual understanding captured by the embeddings. This is illustrated in \cref{fig:s3} for both transformers: \textit{all-MiniLM-L12-v2} and \textit{all-mpnet-base-v2}, where similar patterns are observed compared to when no dimensionality reduction is implemented. The results for UMAP are presented in Appendix \ref{AppendixB1}, \textcolor{blue}{where we observe slightly different patterns}, suggesting that UMAP may still be a better fit for other datasets.

In the next section, we will also analyze how both scenarios impact the predictive model's performance.

\subsection{XGBOOST model performance}\label{XGBpreform}

In all cases, the vectors obtained by the transformers are included as numerical features 
in the XGBOOST model, as this algorithm demonstrated the best performance in the original implementation presented in \cite{papavasileiou2024integrating}. At this point, it is interesting to explore how the amount of information considered during training influences the model's performance. To address this, we will examine two important metrics in supervised machine learning models that provide insight into their accuracy and robustness: mean squared error (MSE) and $R^2$ score.

With this in mind, the next two sections will discuss the results obtained from applying the $k$-fold cross-validation method with $k=10$, while considering progressively increasing portions of the data used for training. 

\subsection{Mean squared error, MSE}\label{Sec:54}

\begin{figure*}[!ht]
	\centering
	\begin{subfigure}[t]{\textwidth}
		\centering
		\includegraphics[width=0.5\linewidth]{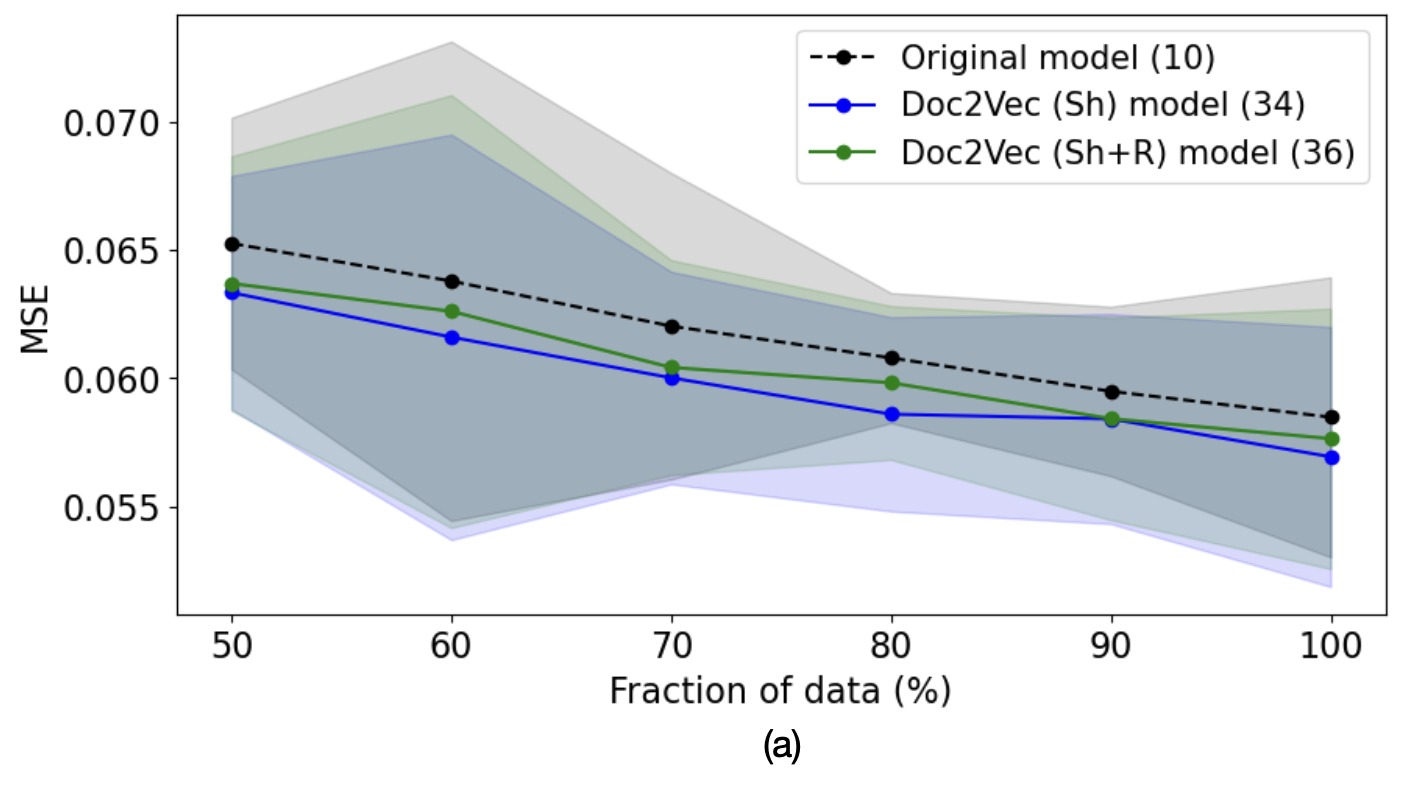}
	\end{subfigure}
	\begin{subfigure}[t]{\textwidth}
		\centering
		\includegraphics[width= \linewidth]{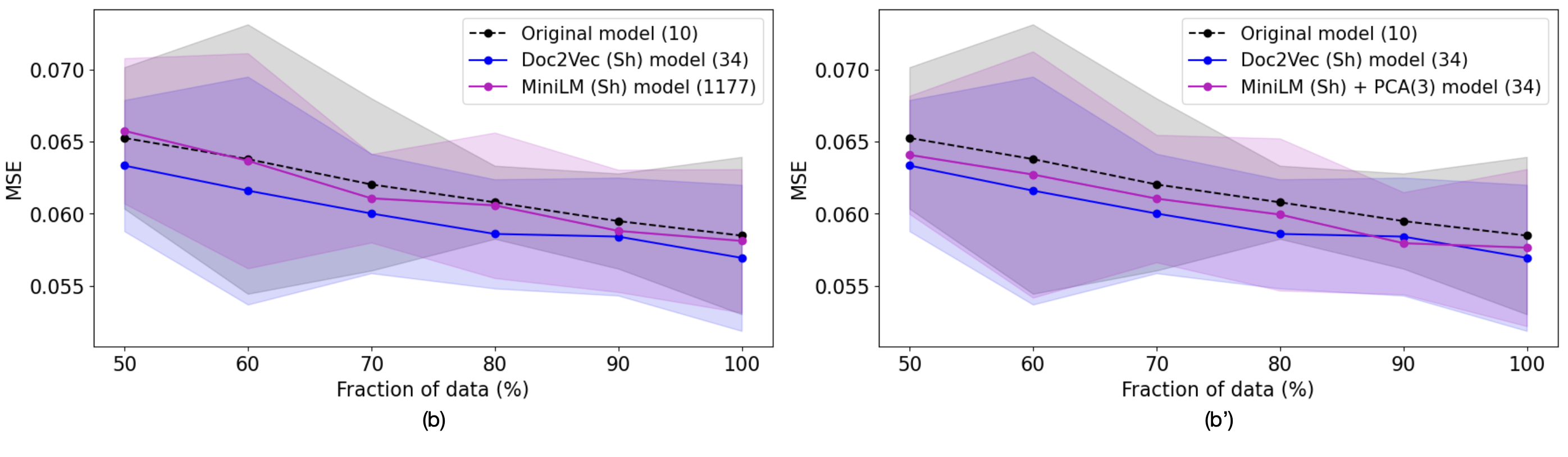}
	\end{subfigure}
	\begin{subfigure}[t]{\textwidth}
		\includegraphics[width= \linewidth]{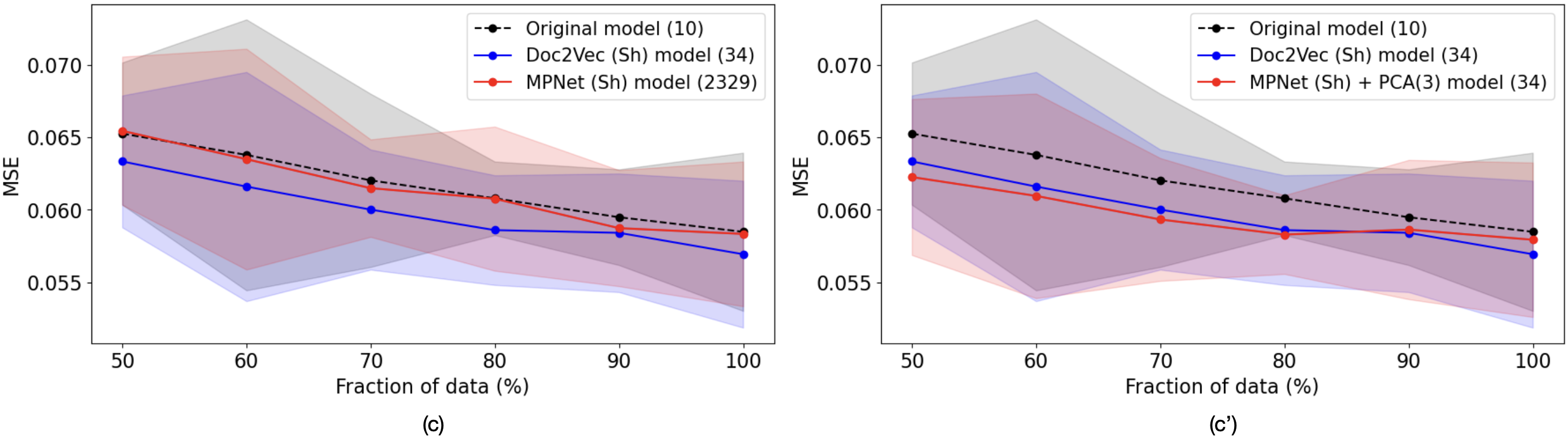}
	\end{subfigure}
	\caption{Confidence intervals for the average MSE score on the test set, computed using a one-standard-deviation bandwidth, are presented after performing 10-fold cross-validation and evaluating progressively increasing fractions of the training data, as indicated on the horizontal axis. (a) Results for the original predictive model (in black), and the predictive model incorporating \textit{Doc2Vec} embeddings for only the \textcolor{Cerulean}{\texttt{Insert shape}} (Sh) feature (in blue), and the predictive model with \textit{Doc2Vec} embeddings for both the \textcolor{Cerulean}{\texttt{Insert shape}} and the production \enquote{recipe} (Sh+R) features (in green). (b) Results for the original model (in black), the model using \textit{Doc2Vec} embeddings (in blue), and the model using \textit{all-MiniLM-L12-v2} embeddings (in purple) for encoding the \textcolor{Cerulean}{\texttt{Insert shape}} (Sh) feature. (b') The same results as in (b), but including \textit{Doc2Vec} embeddings after dimensionality reduction with PCA. (c) Results for the original model (in black), the model using \textit{Doc2Vec} embeddings (in blue), and the model using \textit{all-mpnet-base-v2} embeddings (in red), all applied to encode only the \textcolor{Cerulean}{\texttt{Insert shape}} (Sh) feature. (c') The same results as in (c), but including \textit{Doc2Vec} embeddings after dimensionality reduction using PCA.}
	\label{fig:mse}	
\end{figure*}

In this section, we focus on the first metric, MSE. Results are presented as 99.9 $\%$ confidence intervals for the average MSE, using a bandwidth of one standard deviation (Fig. \ref{fig:mse}). This follows from the fact that the MSE values have a normal distribution (null hypothesis), as shown using the Kolmogorov-Smirnov test in \cref{KSMSE}; the p-values (> 0.05), which represent the probability that the observed values occur under the null hypothesis, in each case allow us to assume the normality of the MSE values.

\begin{center}
\begin{table}[!ht]
\caption{Kolmogorov-Smirnov test for normality (p-values) for MSE values obtained from cross-validation across different models and varying data proportions.\label{KSMSE}}
\centering 
\begin{tabular}{ |c| } 
 \hline
 Fraction \\
 of data (\%) \\ 
 \hline
 50 \\ 
 \hline
 60 \\ 
 \hline
 70 \\ 
 \hline
 80 \\ 
 \hline
 90 \\ 
 \hline
 100 \\ 
 \hline
\end{tabular}
\begin{tabular}{ |c| } 
 \hline
 \textit{Original model} \\
  p-value \\ 
 \hline
 0.77869 \\ 
 \hline
 0.90639 \\ 
 \hline
 0.90014 \\ 
 \hline
 0.42233 \\ 
 \hline
 0.61180 \\ 
 \hline
 0.99849 \\ 
 \hline
\end{tabular}
\begin{tabular}{ |c| } 
 \hline
 \textit{Doc2Vec (Sh)} \\
 p-value \\ 
 \hline 0.95348 \\ 
 \hline
 0.98867 \\ 
 \hline
 0.99588 \\ 
 \hline
 0.97935 \\ 
 \hline
 0.99791 \\ 
 \hline
 0.88989 \\ 
 \hline
\end{tabular}
\begin{tabular}{ |c| } 
 \hline
 \textit{Doc2Vec (Sh+R)} \\
 p-value \\ 
 \hline
 0.91981 \\ 
 \hline
 0.99952 \\ 
 \hline
 0.80851 \\ 
 \hline
 0.99314 \\ 
 \hline
 0.97609 \\ 
 \hline
 0.93013 \\ 
 \hline
\end{tabular}
\begin{tabular}{ |c| } 
 \hline
 \textit{MiniLM (Sh)} \\
  p-value \\ 
 \hline
 0.91337 \\ 
 \hline
 0.99382 \\ 
 \hline
 0.65138 \\ 
 \hline
 0.84190 \\ 
 \hline
 0.45006 \\ 
 \hline
 0.94460 \\ 
 \hline
\end{tabular}
\begin{tabular}{ |c| } 
 \hline
 \textit{MPNet (Sh)} \\
  p-value \\ 
 \hline
 0.80502 \\ 
 \hline
 0.90611 \\ 
 \hline
 0.99869 \\ 
 \hline
 0.75723 \\ 
 \hline
 0.49358 \\ 
 \hline
 0.96711 \\ 
 \hline
\end{tabular}
\end{table}
\end{center}

In \cref{fig:mse} (a), we illustrate the results for the original model, which shows a decreasing linear trend as the training data size increases (black dashed line). However, its confidence interval also narrows after considering 80\% of the training data, suggesting reduced variability around the MSE mean and greater confidence that the true MSE value lies within this (black) interval. Comparing all intervals, this would indicate possible limitations in further reducing the error for the original model. 
When examining the curves for the cases where \textit{Doc2Vec} is used to encode either only the \textcolor{Cerulean}{\texttt{Insert shape}} features (in blue) or both the \textcolor{Cerulean}{\texttt{Insert shape}} features and the recipe (in green), we observe that the average MSE follows a similar decreasing trend. Additionally, we see a narrowing of their respective confidence intervals after considering 70\% of the data, and these intervals cover a slightly lower range than the original model's, suggesting that further reduction in error may be possible, as the bounds of both confidence intervals (blue and green) are slightly below that of the original model. This could also be achieved by incorporating more data or by fine-tuning the model's parameters.


In the other two cases, corresponding to the application of \textit{all-MiniLM-L12-v2} (\cref{fig:mse} (b) and (b')) and \textit{all-mpnet-base-v2} (\cref{fig:mse} (c) and (c'))  transformers with and without the implementation of PCA, respectively), we observe similar behavior to \textit{Doc2Vec}, except that the average error in each case is very close to the average MSE of the original model, indicating no significant improvements. Furthermore, in the case of transformers, considering 80\% of the training data results in a widening of their confidence intervals, suggesting that these representations require as much data as possible to achieve performance equal to or only marginally better than that of the original model. This approach could reduce the variability of the MSE mean, resulting in narrower intervals and providing greater confidence that the true value falls within this range.

\subsection{R$^2$ score}\label{Sec:55}

Similarly to the previous section, we will analyze the evolution of the $R^2$ metric as a function of the proportion of data used for training, through $k$-fold cross-validation. Then, to assess whether the obtained $R^2$ values follow a normal distribution, we again use the Kolmogorov-Smirnov test. As shown in \cref{KSMr2}, the p-values (> 0.05) support the assumption of normality.

\begin{center}
\begin{table}[!ht]
\caption{Kolmogorov-Smirnov test for normality (p-values) for $R^2$ values obtained from cross-validation across different models and varying data proportions. \label{KSMr2}}
\centering 
\begin{tabular}{ |c| } 
 \hline
 Fraction \\
 of data (\%) \\ 
 \hline
 50 \\ 
 \hline
 60 \\ 
 \hline
 70 \\ 
 \hline
 80 \\ 
 \hline
 90 \\ 
 \hline
 100 \\ 
 \hline
\end{tabular}
\begin{tabular}{ |c| } 
 \hline
 \textit{Original model} \\
  p-value \\ 
 \hline
 0.53053 \\ 
 \hline
 0.30654 \\ 
 \hline
 0.49133 \\ 
 \hline
 0.57835 \\ 
 \hline
 0.96425 \\ 
 \hline
 0.74970 \\ 
 \hline
\end{tabular}
\begin{tabular}{ |c| } 
 \hline
 \textit{Doc2Vec (Sh)} \\
 p-value \\ 
 \hline 
 0.51809 \\ 
 \hline
 0.83924 \\ 
 \hline
 0.41014 \\ 
 \hline
 0.47981 \\ 
 \hline
 0.99058 \\ 
 \hline
 0.83713 \\ 
 \hline
\end{tabular}
\begin{tabular}{ |c| } 
 \hline
 \textit{Doc2Vec (Sh+R)} \\
 p-value \\ 
 \hline
 0.74102 \\ 
 \hline
 0.92346 \\ 
 \hline
 0.41024 \\ 
 \hline
 0.62140 \\ 
 \hline
 0.98585 \\ 
 \hline
 0.71981 \\ 
 \hline
\end{tabular}
\begin{tabular}{ |c| } 
 \hline
 \textit{MiniLM (Sh)} \\
  p-value \\ 
 \hline
 0.72533 \\ 
 \hline
 0.83543 \\ 
 \hline
 0.32626 \\ 
 \hline
 0.42428 \\ 
 \hline
 0.97045 \\ 
 \hline
 0.74681 \\ 
 \hline
\end{tabular}
\begin{tabular}{ |c| } 
 \hline
 \textit{MPNet (Sh)} \\
  p-value \\ 
 \hline
 0.76436 \\ 
 \hline
 0.69900 \\ 
 \hline
 0.47973 \\ 
 \hline
 0.49954 \\ 
 \hline
 0.98269 \\ 
 \hline
 0.76401 \\ 
 \hline
\end{tabular}
\end{table}
\end{center}

\begin{figure}[!ht]
	\centering
	\begin{subfigure}[t]{\textwidth}
		\centering
		\includegraphics[width=0.5\linewidth]{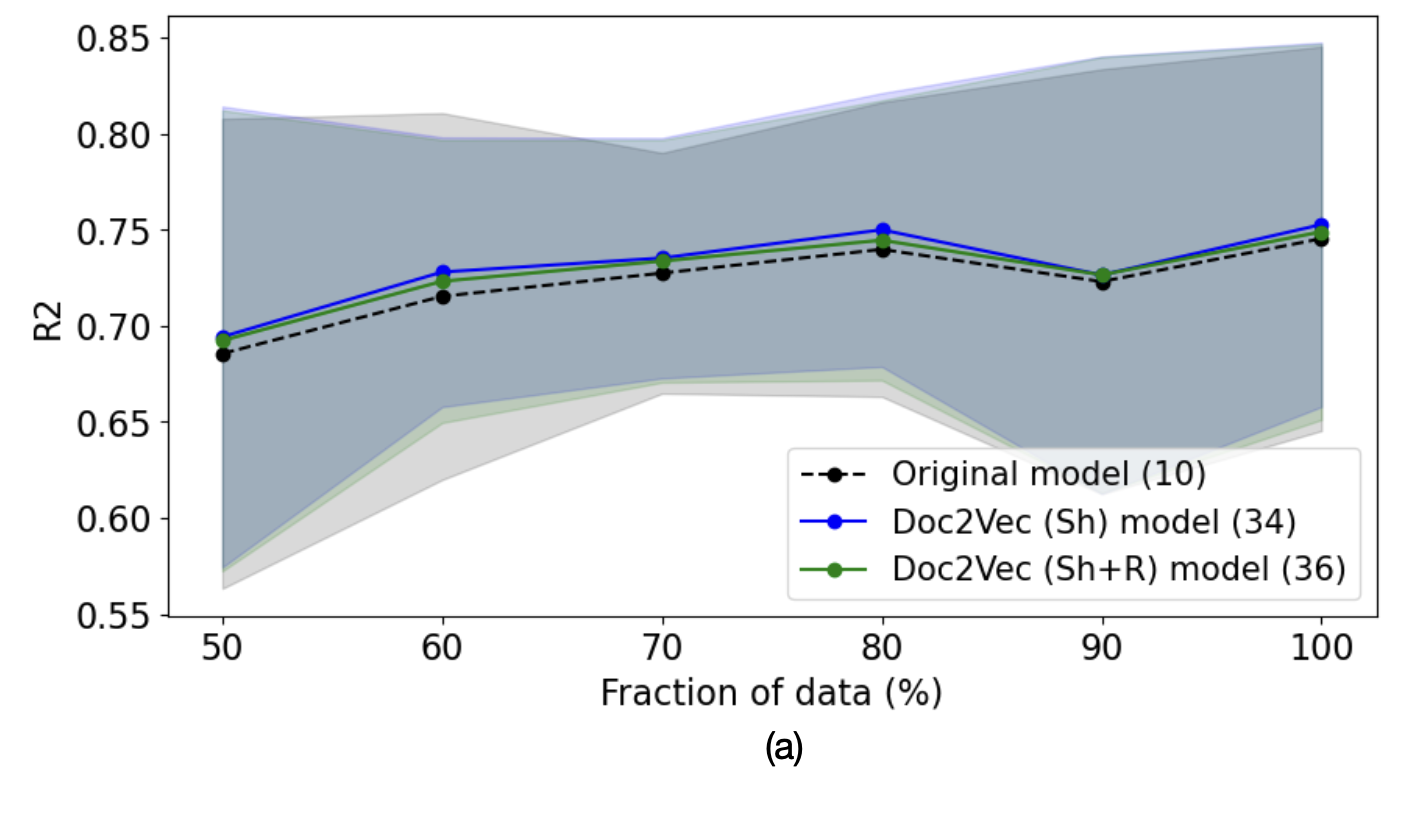}
	\end{subfigure}
	\begin{subfigure}[t]{\textwidth}
		\centering
		\includegraphics[width= \linewidth]{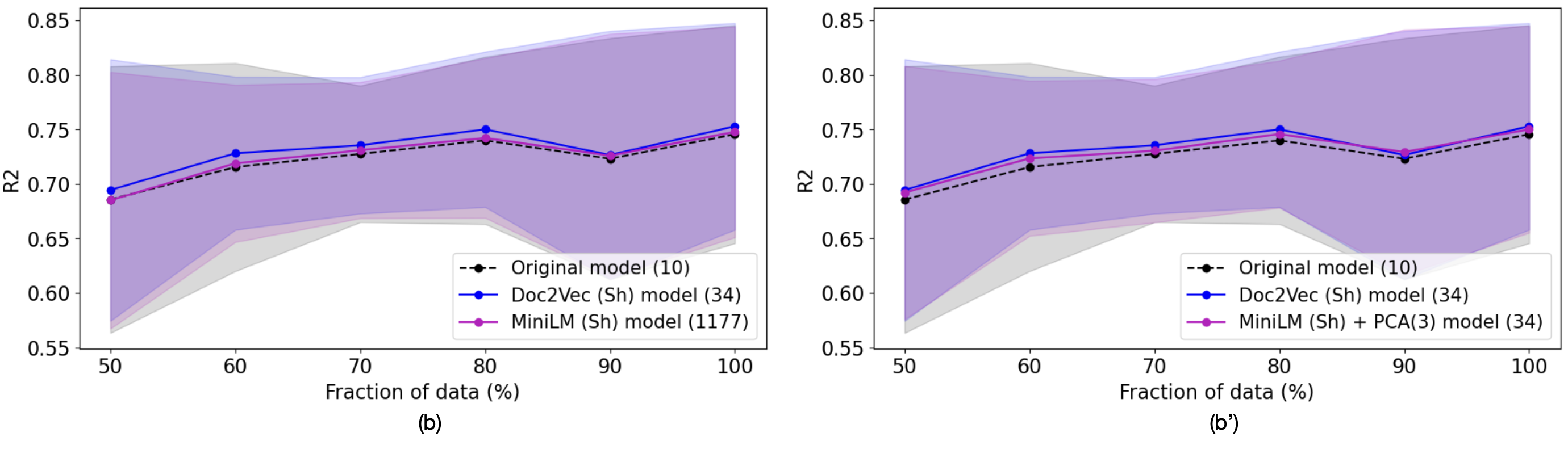}
	\end{subfigure}
	\begin{subfigure}[t]{\textwidth}
		\includegraphics[width= \linewidth]{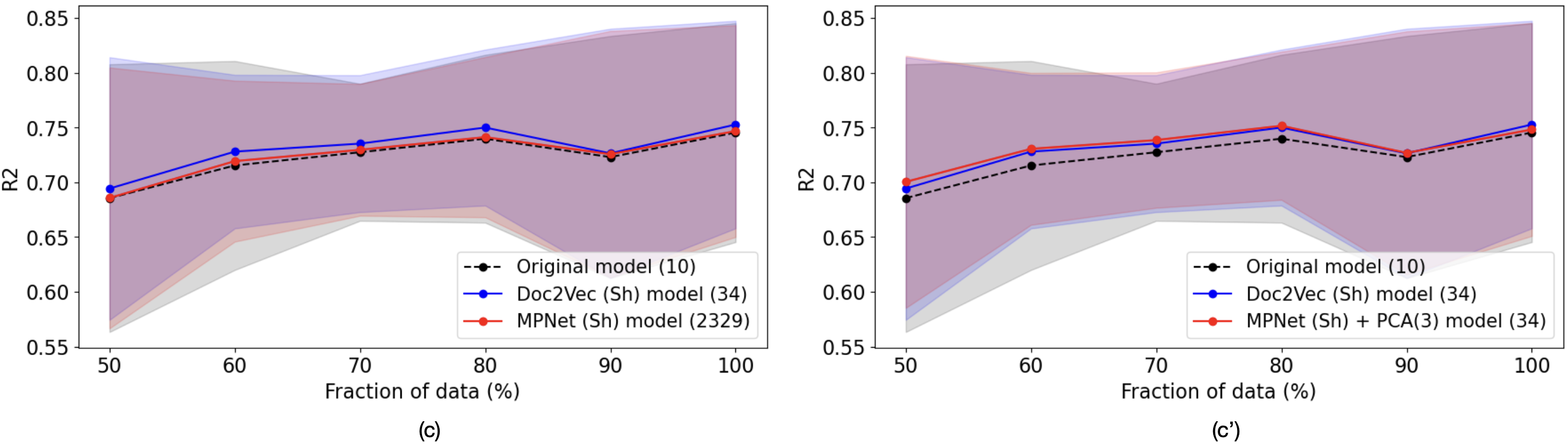}
	\end{subfigure}
	\caption{Confidence intervals for the average $R^2$ score on the test set, computed using a one-standard-deviation bandwidth, are presented after performing 10-fold cross-validation and evaluating progressively increasing fractions of the training data, as indicated on the horizontal axis. (a) Results for the original predictive model (in black), and the predictive model including \textit{Doc2Vec} embeddings for only the \textcolor{Cerulean}{\texttt{Insert shape}} (Sh) feature (in blue), and the \textit{Doc2Vec} embeddings for both the \textcolor{Cerulean}{\texttt{Insert shape}} and the production \enquote{recipe} (Sh+R) features (in green). (b) Results for the original model (in black), \textit{Doc2Vec} embeddings (in blue) and \textit{all-MiniLM-L12-v2} embeddings (in purple), used for encoding only the \textcolor{Cerulean}{\texttt{Insert shape}} (Sh). (b') The same results as in (b) but including \textit{Doc2Vec} embeddings after dimensionality reduction using PCA. (c) Results for the original model (in black), \textit{Doc2Vec} embeddings (in blue) and \textit{all-mpnet-base-v2} embeddings (in red), also used for encoding only the \textcolor{Cerulean}{\texttt{Insert shape}} features (Sh). (c') The same results as in (c) but including \textit{Doc2Vec} embeddings after dimensionality reduction with PCA.} \label{fig:r2}
\end{figure}

Thus, in Fig. \ref{fig:r2}, we explore the original model performance guides by the $R^2$ score, which shows an increasing trend as the training data size grows (dashed black line). Its confidence interval also widens downward once 90\% of the training data is considered, which means it limits the potential for improvement in this metric, as the bounds of the other (blue and green) confidence intervals within which the true value of $R^2$ is likely to fall are slightly above those of the original model. 

In contrast, when examining the curves for models using \textit{Doc2Vec} to encode only the \textcolor{Cerulean}{\texttt{Insert shape}} features (in blue) or both the shapes of the inserts and the recipe (in green), the average $R^2$ follows a similar increasing trend, slightly above the black curve. Additionally, their respective confidence intervals also widen once 90\% of the data is included.

In the remaining two cases, corresponding to the implementation of \textit{all-MiniLM-L12-v2} (\cref{fig:r2} (b) and (b')) and \textit{all-mpnet-base-v2} (\cref{fig:r2} (c) and (c'))  transformers with and without the implementation of PCA, respectively), similar behaviors are observed, although the average scores in each case are closer to the original model's average $R^2$, indicating no significant improvements. For transformers, the same widening of confidence intervals is noted when considering 90\% of the training data, suggesting that the model requires as much data as possible to achieve only slight improvements in performance. Again, this approach helps to reduce the variability of the $R^2$ mean, ensuring that the true value falls within this interval.

Furthermore, parameter optimization was performed on the XGBOOST regression parameters for each of the models used for encoding categorical variables. In all cases, the optimized parameters were found to be close to those originally set. Then, a similar analysis, including 10-fold cross-validation with varying portions of the data for training, yielded results consistent with those presented in the previous sections.

Finally, to analyze the performance of the XGBOOST model when varying the dimensionality of the latent spaces generated by transformers and embedding, we also estimated the confidence intervals for the average MSE and $R^2$. Considering much larger latent space dimensions for the \textit{Doc2Vec} model do not show significant improvements compared to those already achieved with a dimension of 3. However, it is more interesting to observe the effects when reducing the latent space dimension of the transformers to 3 (cf. Figs. \ref{fig:mse} (b') and (c'), \ref{fig:r2} (b') and (c'), \ref{fig:r2UMAP}, and \ref{fig:MSEUMAP}). The main advantages here are a slight improvement in metrics and increased efficiency of the predictive model, as it has to handle fewer variables.\\

Another important aspect to mention at this point is the computational cost of each of the models presented. First, it is worth noting that there is no significant computational cost associated with training the \textit{Doc2Vec} embedding due to the small quantity and short length of the descriptions, as well as the fact that the latent space dimension, set to 3, is low. Additionally, the computational cost of applying the two transformers (\textit{all-MiniLM-L12-v2} and \textit{all-MPNet-base-v2}) is negligible because both are pre-trained models and their respective encoding speeds are highly efficient, as reported in \Cref{speed} (cf. \cite{huggingfaces}).

\begin{table}[ht]
	\begin{center}
		\caption{Computational cost of embeddings and transformers: model parameters and encoding speed on a V100 GPU.}
			\begin{tabular}{ccc}
				\hline
				\textbf{Model} & \textbf{\# Parameters} & \textbf{Encoding speed} \\
				 &  & \textbf{(sentences/sec)} \\
				\hline
				Doc2Vec & $\sim 50$ for this application & 50K$-$100K \\
			\textit{all-MiniLM-L12-v2} & 33.4M & 7.5K\\
				\textit{all-mpnet-base-v2} & 109M & 2.8K \\
				\hline
			\end{tabular}\label{speed}
	\end{center}
\end{table}

The computational cost arises when training the predictive model, which incorporates the vector representations of the insert shapes descriptions using embedding and transformers. Each model is trained six times, once for each dataset proportion (see \cref{fig:mse} and \cref{fig:r2}). Thus, each of the predictive models using the vector representation obtained through the \textit{Doc2Vec} embedding takes, on average, 23 seconds per run. In contrast, the predictive models that include high-dimensional vector representations produced by the transformers have a much higher computational cost. The \textit{all-MiniLM-L12-v2} model takes, on average, about 10 minutes per run, while the \textit{all-mpnet-base-v2} model can take approximately 20 minutes per run. This demonstrates a clear correlation between dimensionality and the computational cost of each model.

However, this problem caused by high dimensionality can be successfully addressed by reducing the dimensionality of the vector representations with methods such as PCA or UMAP, as detailed in \Cref{sec:Similarity} and shown in \Cref{Sec:54}, \Cref{Sec:55} and Appendix \ref{Appendix}. This approach yields results with high precision and an average runtime of around 25 seconds, thus recovering efficiency without losing accuracy.

\subsection{Feature importance and Shapley Analysis}

Recall that for the feature importance with \texttt{total\_gain} is based on the total reduction in error achieved by splitting all the trees, indicating how much each input feature contributes to reducing the overall error of the XGBOOST model. Similarly, the Shapley analysis focuses on each feature contribution to the model predictions. In this context, we will analyze how the implementation of transformers for the encoding of categorical features affects the corresponding importance scores for the input features. 

Additionally, it is important to mention that these values do not follow to a standardized range, and there is no upper limit to the \texttt{total\_gain} score or the Shapley scores, as we are summing the overall contributions of the feature. As a result, the \texttt{total\_gain} score can potentially increase indefinitely, reflecting the cumulative impact of all contributing factors.

\begin{figure}[!ht]
	\centering
	\includegraphics[width=.8\linewidth]{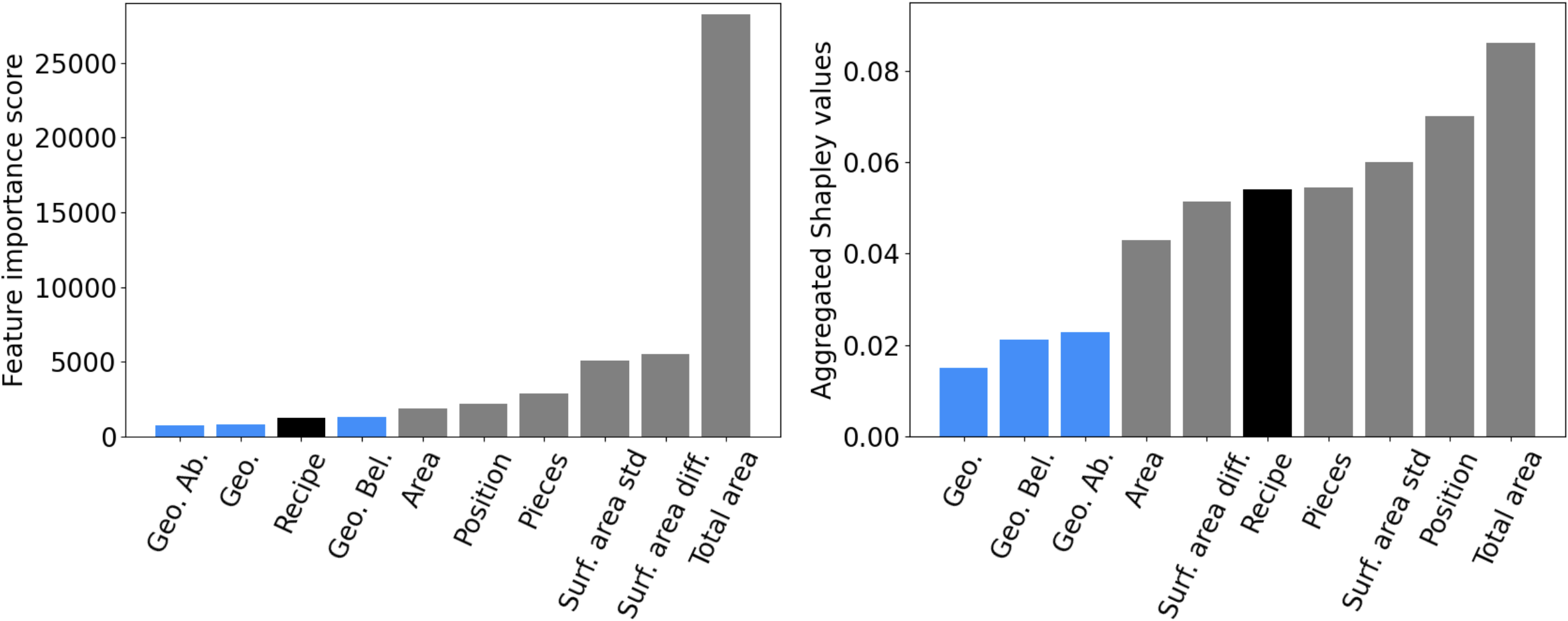}
	\caption{For the original model, from left to right: the first figure shows the top ten input features with the highest scores for the XGBOOST model, using feature importance measured by \texttt{total\_gain}. The second figure displays the top ten input features with the highest weighted scores obtained after performing Shapley analysis. In both cases, the gray bars represent numerical features, the black bars represent the categorical feature called "recipe", and the blue bars correspond to categorical variables indicating the geometries of the inserts located at the current position, above, and below.}
	\label{fig:IMPp}
\end{figure}

For the original model (see \cref{fig:IMPp}), which uses binary encoding for handling categorical features, we observe that numerical features (in gray) are the most influential in terms of feature importance scores, particularly those related to total area and surface variability. In contrast, categorical variables related to insert geometries (in blue) and the recipe (in black) have considerably lower scores. Shapley analysis shows a similar trend where numerical features have higher scores than categorical ones. This suggests that potentially critical inputs for the process are neglected as categorical variables.


Based on subject matter expertise, we recognize the significance of categorical variables; however, quantifying this importance directly in the original model is not possible. With the implementation of embeddings, we can now effectively quantify feature importance. Furthermore, we observe that categorical features, after being transformed through embeddings, have begun to dominate and become increasingly significant.

\begin{figure}[!ht]
	\centering
	\includegraphics[width=.8\linewidth]{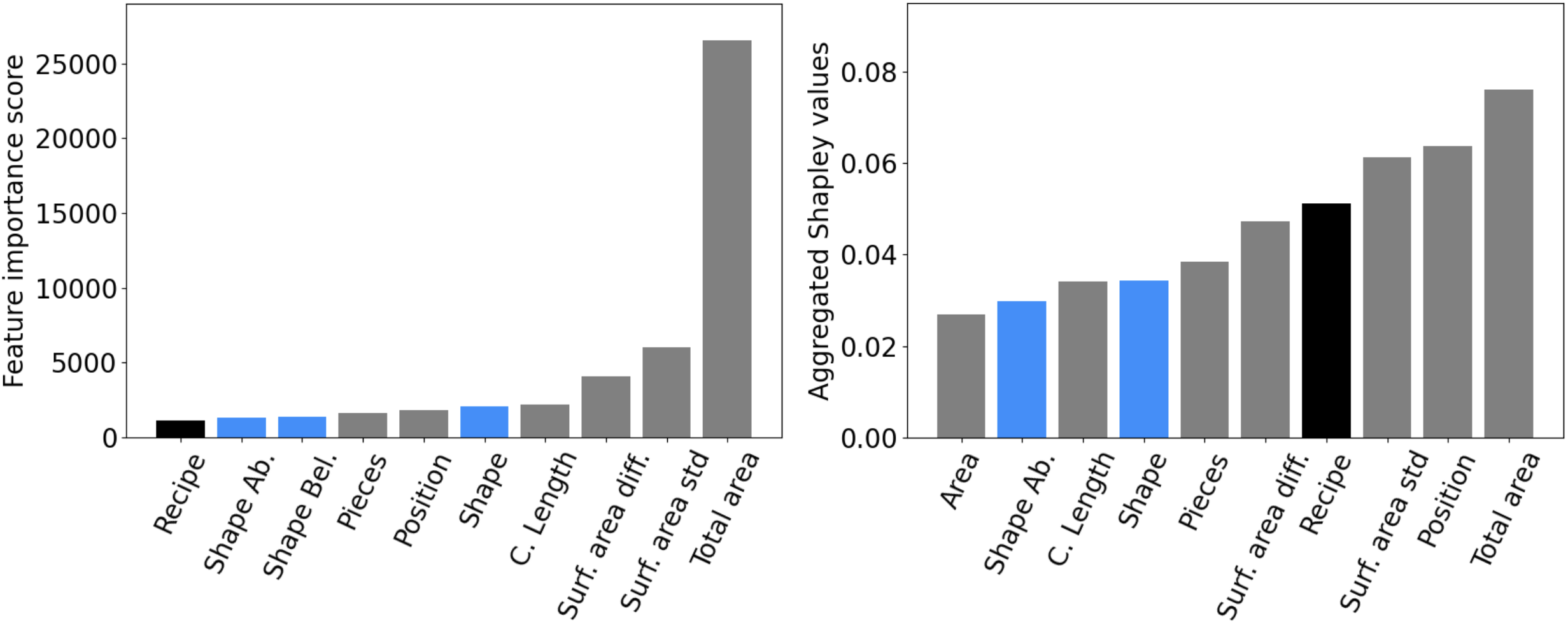}
	\caption{For the \textit{Doc2Vec} model used to encode only the \textcolor{Cerulean}{\texttt{Insert shape}} features, from left to right: the first figure shows the top ten input features with the highest scores for the XGBOOST model, based on \texttt{total\_gain} feature importance. Following this, the next figure displays the top ten input features with the highest weighted scores obtained from Shapley analysis. In both cases, the gray bars represent numerical features, the black bars correspond to the categorical feature called "recipe", and the blue bars indicate categorical variables related to the insert shapes at the current, above and below positions.}
	\label{fig:IMPd2vs}
\end{figure}

Both feature importance and Shapley analysis suggest that, although some numerical features remain the most important, we observe a slight increase in the importance of insert shapes after being encoded with \textit{Doc2Vec} (see \cref{fig:IMPd2vs}). It is also important to note that the number of input features has tripled following preprocessing and encoding. This indicates that incorporating more useful information into the predictive model is indeed possible by transforming the descriptions of insert shapes using an embedding.

\begin{figure}[!ht]
	\centering
	\includegraphics[width=.8\linewidth]{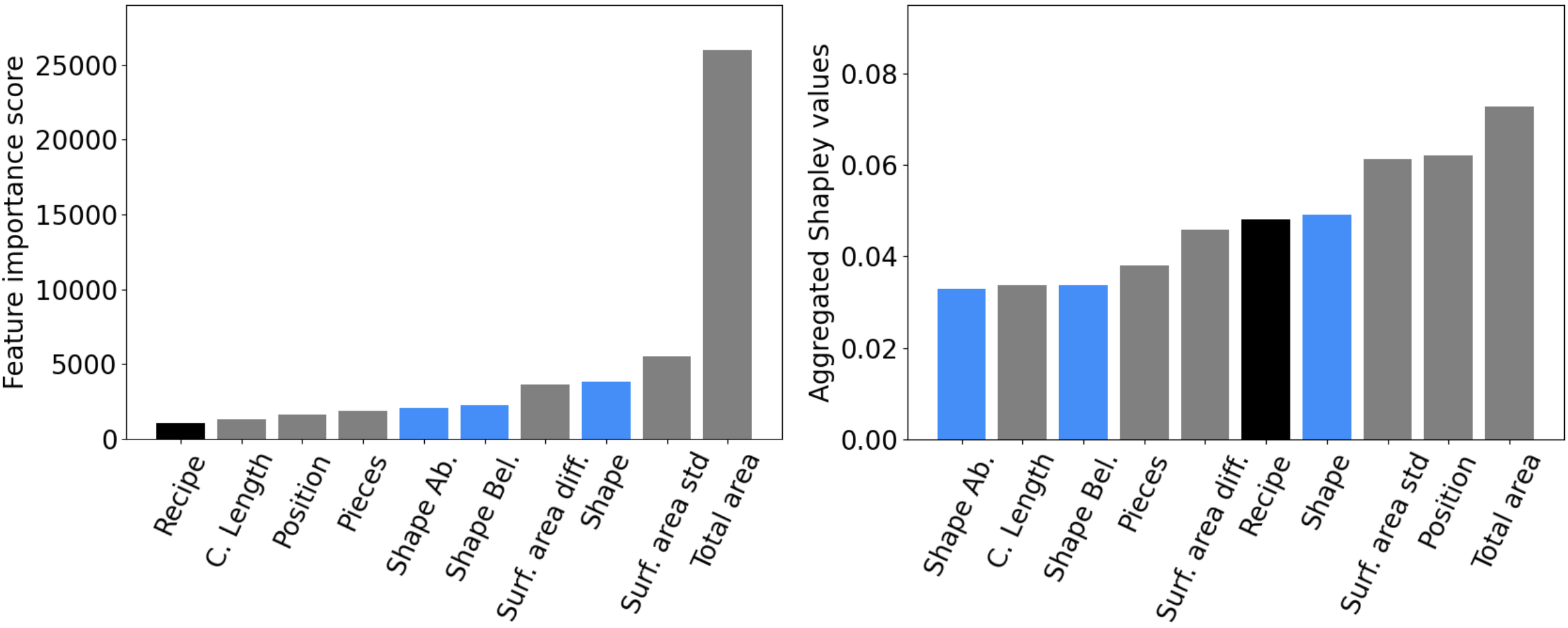}
	\caption{For the \textit{all-MiniLM-L12-v2} model that encodes only the \textcolor{Cerulean}{\texttt{Insert shape}}   features, from left to right: the first figure presents the top ten input features with the highest scores for the XGBOOST model, based on \texttt{total\_gain} feature importance. The subsequent figure shows the top ten input features with the highest weighted scores derived from Shapley analysis. In both cases, the gray bars represent numerical features, while the black bars correspond to the categorical feature called "recipe". Finally, the blue bars indicate categorical variables related to the insert shapes located at the current position, above, and below.}
	\label{fig:IMPml}
\end{figure}

\begin{figure}[!ht]
	\centering
	\includegraphics[width=.8\linewidth]{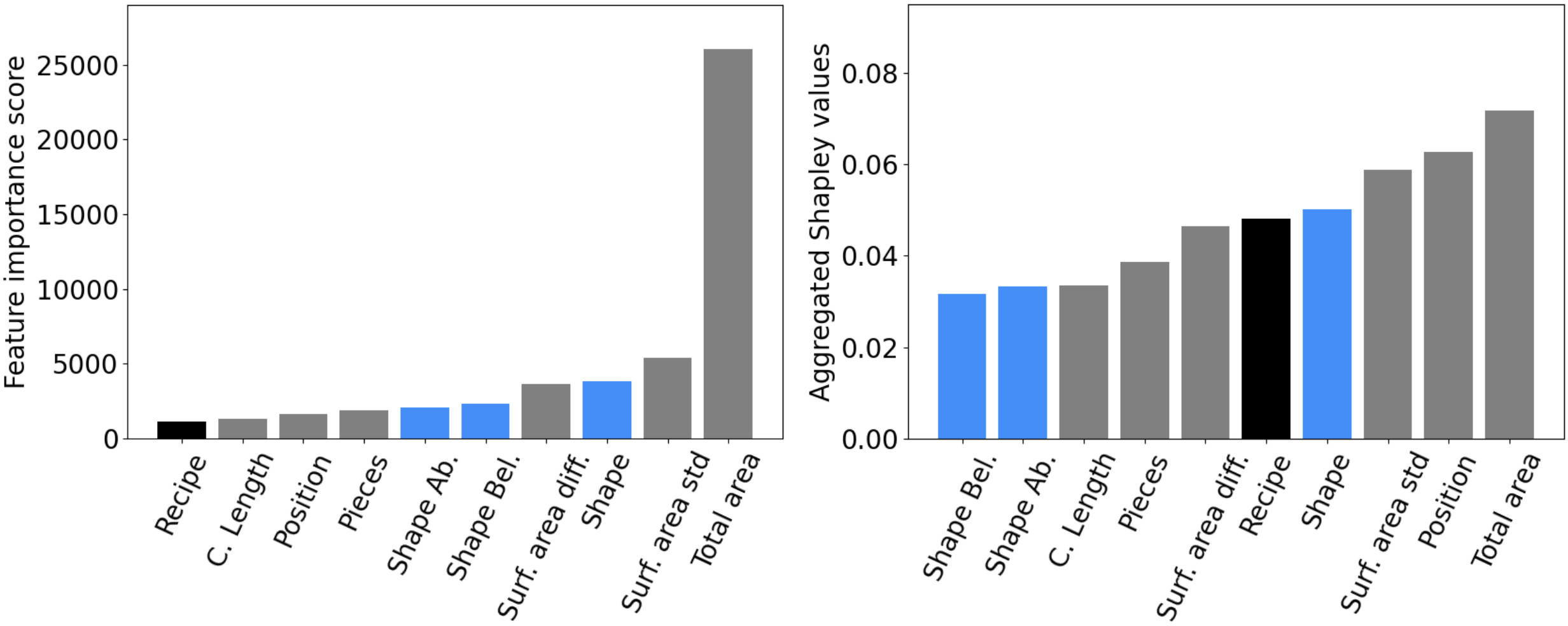}
	\caption{For the \textit{all-mpnet-base-v2} model used to encode only the \textcolor{Cerulean}{\texttt{Insert shape}} features, the figures are presented as follows from left to right: the first figure illustrates the top ten input features with the highest scores for the XGBOOST model, based on \texttt{total\_gain} feature importance. Then, the second figure shows the top ten input features with the highest weighted scores as determined by Shapley analysis. In both cases, the gray bars represent numerical features, the black bars indicate the categorical feature called "recipe", and the blue bars correspond to categorical variables representing the insert shapes located at the current position, as well as those above and below it.}
	\label{fig:IMPnp}
\end{figure}

The use of more sophisticated models as transformers to encode categorical features (see \cref{fig:IMPml} and \cref{fig:IMPnp}), show a similar trend in terms of feature importance and Shapley analysis: categorical variables related to the shapes of the inserts are ranked within the top 5. This further emphasizes the value of having interpreted textual descriptions more robustly for inclusion in the predictive model.

Finally, with this set of tools, we can initiate the optimization of the reactor setup. An initial step involves leveraging the above results to identify the most influential input parameters for the predictive model. This analysis can be further refined through sensitivity analysis and uncertainty quantification, enabling a deeper understanding of how these parameters contribute to process variability. By systematically addressing these factors, we aim to mitigate variability and enhance process stability. This effort remains an integral part of our ongoing research.

\section{Conclusions}
This study highlights the benefits of leveraging transformer-based embeddings for processing categorical variables, particularly in industrial applications. The case study presented here, concerns the industrial-scale Chemical Vapour Deposition of coatings, involving actual production data. The nature of the inputs that include both numerical and categorical industrial production data, makes this a suitable example to showcase the performance of the natural language processing models. Nevertheless the findings of this work are not limited to this specific application but can be useful in other cases where there is variety in the type of data involved. Unlike traditional encoding techniques such as one-hot or binary encoding, transformers capture intricate patterns and relationships within categorical data, enabling representations that preserve both semantic meaning and contextual relevance. 

By embedding categorical data in a contextually rich manner, models gain deeper insights, leading to a more comprehensive understanding of critical process parameters. This approach is especially valuable in complex industrial systems, where categorical inputs play a pivotal role in determining outcomes. 

While this study focuses on the methodological advantages of transformer-based embeddings, future research will explore their direct applications in reactor optimization, process efficiency improvements, and the development of novel operational strategies. These extensions present distinct research challenges and merit a separate investigation to ensure a thorough and application-driven analysis.  

Furthermore, this work lays the foundation for incorporating sensitivity analysis and uncertainty quantification into model evaluation. Future studies could explore probabilistic frameworks such as Bayesian neural networks, variational inference, and probabilistic graphical models to enhance both accuracy and interpretability. By integrating these techniques with transformer-based embeddings, future research can refine sensitivity analysis for complex, high-dimensional data, ultimately improving decision-making in industrial applications.  

By demonstrating the advantages of embedding categorical variables using transformers, this work contributes to a broader understanding of how NLP-based techniques can be applied in industrial machine learning. The insights gained serve as a foundation for further exploration, paving the way for more robust, interpretable, and application-driven advancements in process modeling and optimization.

\section*{Acknowledgments}
This research was funded in part by the Luxembourg National Research Fund (FNR), grant reference [16758846]. For the purpose of open access, the authors have applied a Creative Commons Attribution 4.0 International (CC BY 4.0) license to any Author Accepted Manuscript version arising from this submission. PP gratefully acknowledges funding from the FSTM in the University of Luxembourg. 



\bibliographystyle{elsarticle-num-names} 
\bibliography{references}

\newpage
\begin{appendices}
	\crefalias{section}{appsec}
	\section{Appendix}\label{Appendix}
	
    \subsection{Similarity after dimensionality reduction using UMAP}\label{AppendixB1}
	\begin{figure}[!h]
		\centering
		\includegraphics[width=0.65\linewidth]{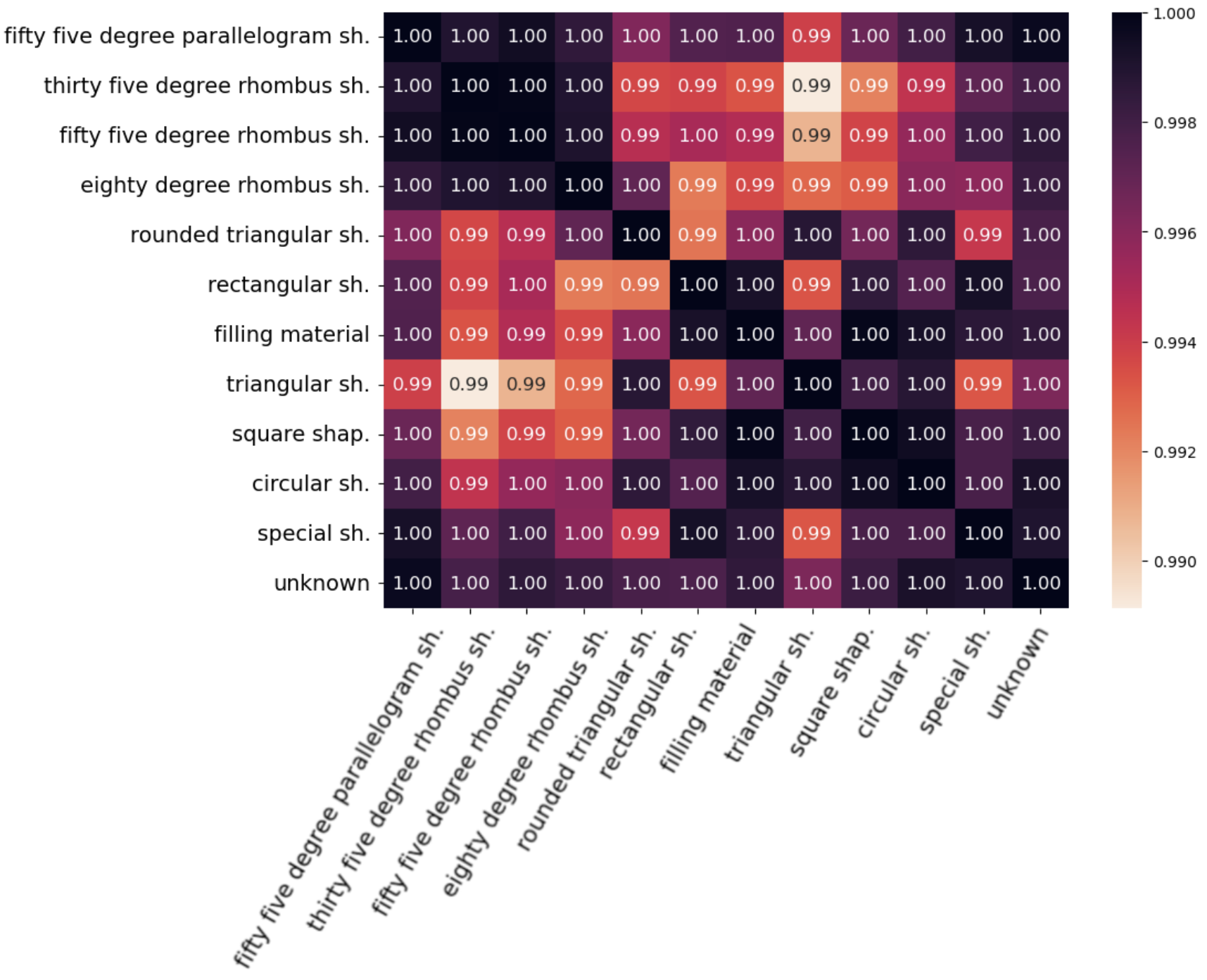}
		\includegraphics[width=0.65\linewidth]{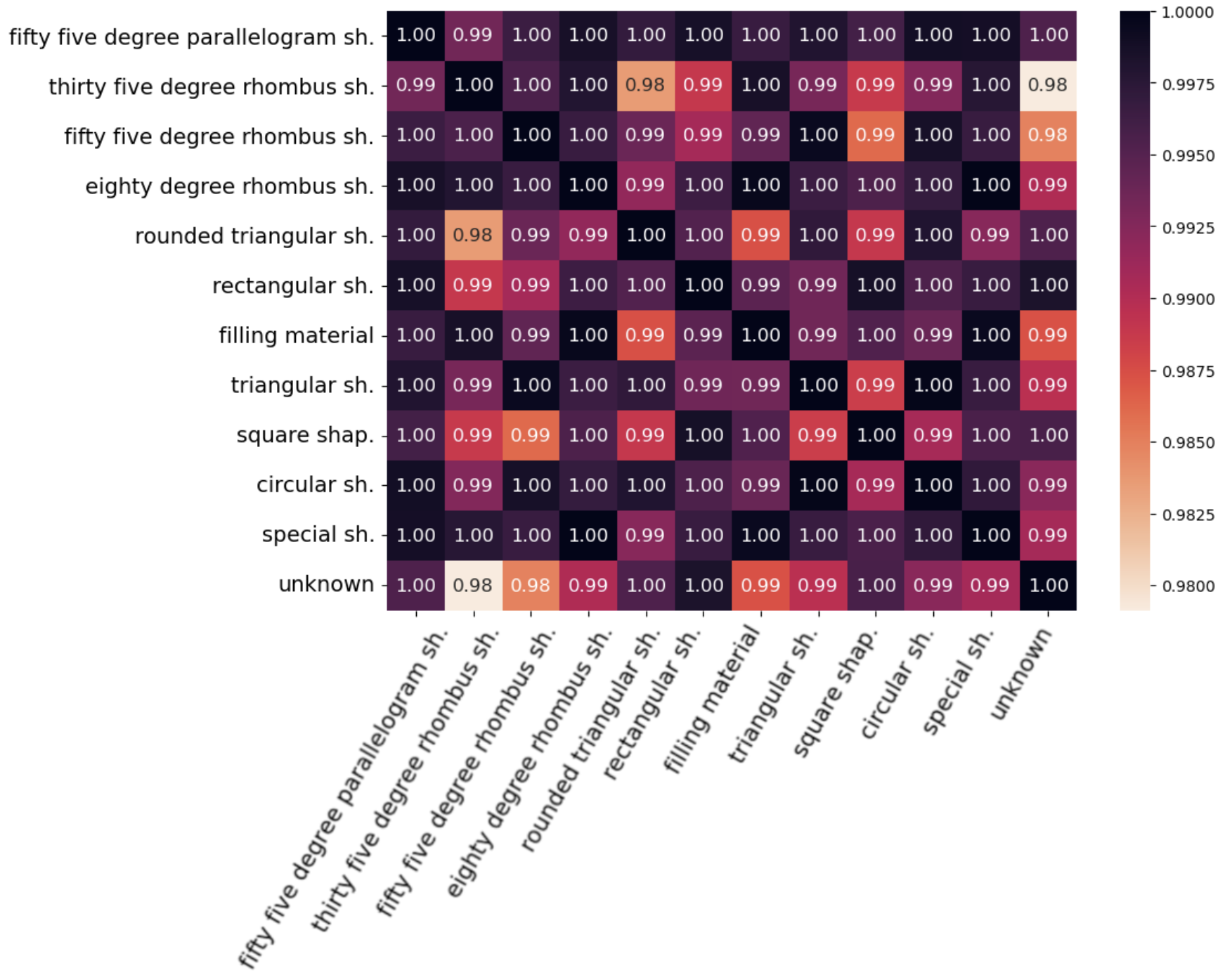}
		\caption{Heatmaps showing cosine similarity values derived for each of the dense vectors generated by the \textit{all-MiniLM-L12-v22} model (top) and  the \textit{all-mpnet-base-v2} model (bottom), both processed through UMAP for dimensionality reduction. The values range from -1 to 1, where lighter shades correspond to lower similarity and darker shades to higher similarity. Shortened versions of the twelve embedded descriptions are used as references on both the horizontal and vertical axes.}
		\label{fig:UMAP}
	\end{figure}

 \newpage
	\subsection{XGBOOST performance after dimensionality reduction using UMAP}\label{AppendixB2}
	\begin{figure}[!h]
		\centering
		\includegraphics[width=0.6\linewidth]{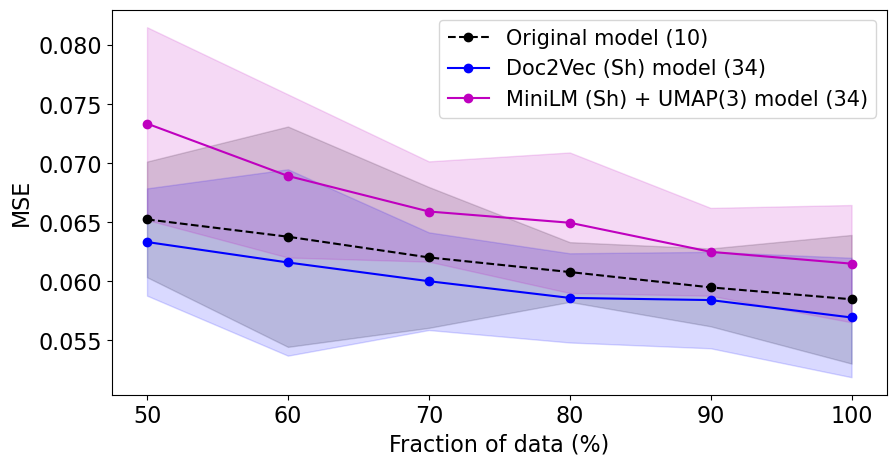}
		\includegraphics[width=0.6\linewidth]{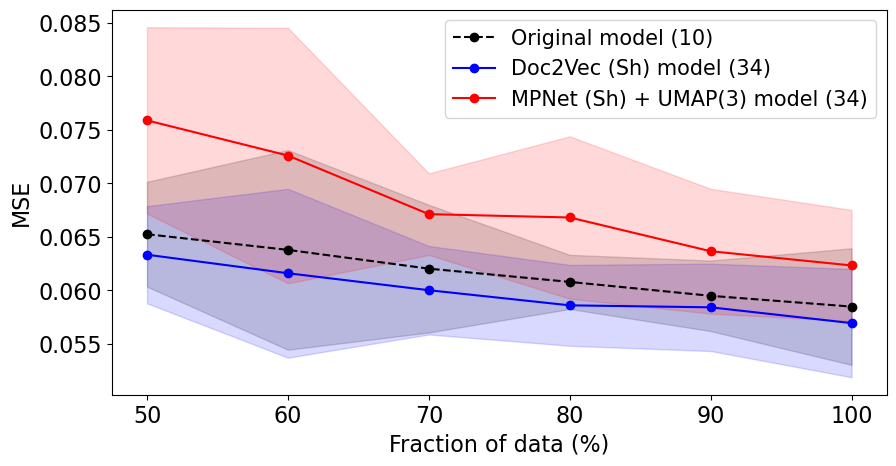}
		\caption{Confidence intervals for the average MSE score on the test set, using a one standard deviation bandwidth, are displayed after performing 10-fold cross-validation and considering progressively increasing fractions of the training data, as shown on the horizontal axis. From top to bottom: the first figure illustrates the confidence intervals for the original model (in black), \textit{Doc2Vec} (in blue), and \textit{all-MiniLM-L12-v2} (in purple), all used for encoding just the \textcolor{Cerulean}{\texttt{Insert shape}} (Sh) and after dimensionality reduction using UMAP. The second figure contrasts the confidence intervals for the original model (in black), \textit{Doc2Vec} (in blue), and \textit{all-mpnet-base-v2} (in red), also used for encoding only the \textcolor{Cerulean}{\texttt{Insert shape}} features (Sh) and after dimensionality reduction using UMAP.} 
         \label{fig:r2UMAP}
	\end{figure}

 \newpage
	\begin{figure}[!h]
		\centering
		\includegraphics[width=0.6\linewidth]{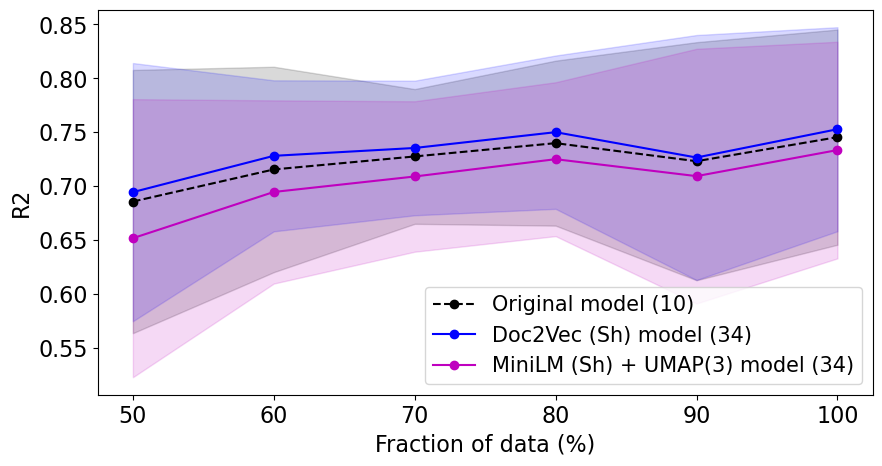}
		\includegraphics[width=0.6\linewidth]{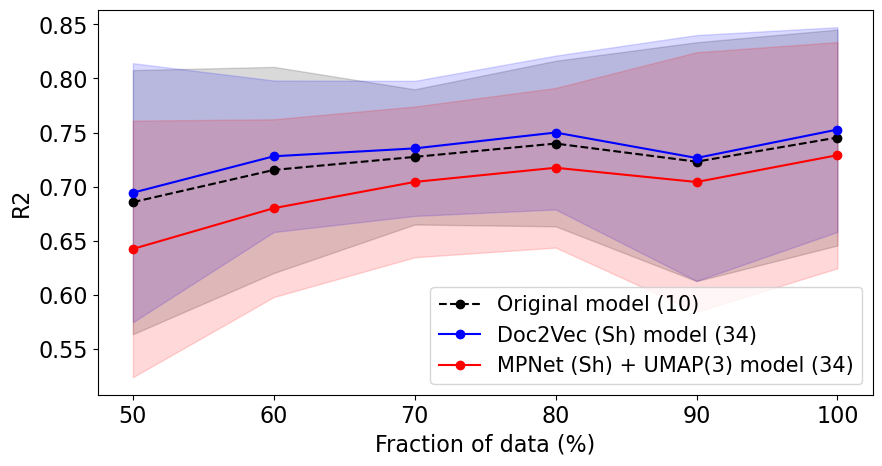}
		\caption{Confidence intervals for the average $R^2$ score on the test set, using a one standard deviation bandwidth, are displayed after performing 10-fold cross-validation and considering progressively increasing fractions of the training data, as shown on the horizontal axis. From top to bottom: the first figure illustrates the intervals for the original model (in black), \textit{Doc2Vec} (in blue), and \textit{all-MiniLM-L12-v2} (in purple), all used for encoding just the \textcolor{Cerulean}{\texttt{Insert shape}} (Sh) and after dimensionality reduction using UMAP. The final figure contrasts the confidence intervals for the original model (in black), \textit{Doc2Vec} (in blue), and \textit{all-mpnet-base-v2} (in red), also used for encoding only the \textcolor{Cerulean}{\texttt{Insert shape}} features (Sh) and after dimensionality reduction using UMAP.} \label{fig:MSEUMAP}
	\end{figure}
\end{appendices}

	
	
\end{document}